\documentclass[10pt,twocolumn,letterpaper]{article}

\usepackage{iccv}
\usepackage{times}
\usepackage{epsfig}
\usepackage{graphicx}
\usepackage{amsmath}
\usepackage{amssymb}

\usepackage{multirow}
\usepackage{booktabs}
\usepackage{array}
\usepackage{makecell}
\usepackage{graphbox}

\usepackage{algorithm} 
\usepackage{algpseudocode}
\usepackage{xcolor}
\usepackage{xspace}

\newcolumntype{P}[1]{>{\centering\arraybackslash}p{#1}}
\newcolumntype{M}[1]{>{\centering\arraybackslash}m{#1}}

\newcommand\st{\textsuperscript{st}\xspace}
\newcommand\nd{\textsuperscript{nd}\xspace}
\newcommand\rd{\textsuperscript{rd}\xspace}
\newcommand\nth{\textsuperscript{th}\xspace}

\DeclareMathOperator*{\argmax}{argmax}

\usepackage[breaklinks=true,bookmarks=false]{hyperref}
\usepackage[accsupp]{axessibility}  

\usepackage[capitalise, noabbrev]{cleveref}
\usepackage{flushend}

\iccvfinalcopy 

\def\iccvPaperID{4584} 

\ificcvfinal\fi

\begin{document}

\title{Proxy Anchor-based Unsupervised Learning for Continuous Generalized Category Discovery}

\author{
	Hyungmin Kim$^{1,2}$\and Sungho Suh$^{3,4}$\and Daehwan Kim$^{2}$\and Daun Jeong$^{2}$\and Hansang Cho$^{2}$\and Junmo Kim$^{1}$ \\
	$^1$Korea Advanced Institute of Science and Technology, Daejeon, South Korea\\
	$^2$Samsung Electro-Mechanics, Suwon, South Korea\\
	$^3$German Research Center for Artificial Intelligence (DFKI), Kaiserslautern, Germany\\
        $^4$Department of Computer Science, RPTU Kaiserslautern-Landau, Kaiserslautern, Germany\\
	{\tt\small \{hyungmin83, junmo.kim\}@kaist.ac.kr, sungho.suh@dfki.de} \\
    {\tt\small \{daehwan85.kim, du33.jeong, hansang.cho\}@samsung.com}
}

\maketitle
\ificcvfinal\thispagestyle{empty}\fi

\begin{abstract}
Recent advances in deep learning have significantly improved the performance of various computer vision applications. However, discovering novel categories in an incremental learning scenario remains a challenging problem due to the lack of prior knowledge about the number and nature of new categories. Existing methods for novel category discovery are limited by their reliance on labeled datasets and prior knowledge about the number of novel categories and the proportion of novel samples in the batch. To address the limitations and more accurately reflect real-world scenarios, in this paper, we propose a novel unsupervised class incremental learning approach for discovering novel categories on unlabeled sets without prior knowledge. The proposed method fine-tunes the feature extractor and proxy anchors on labeled sets, then splits samples into old and novel categories and clusters on the unlabeled dataset. Furthermore, the proxy anchors-based exemplar generates representative category vectors to mitigate catastrophic forgetting. Experimental results demonstrate that our proposed approach outperforms the state-of-the-art methods on fine-grained datasets under real-world scenarios.
\end{abstract}

\vspace{-5mm}
\section{Introduction}
\begin{figure*}[!t]
    \centering
    \resizebox{1.\linewidth}{!}{
    \setlength{\tabcolsep}{1pt}
    \begin{tabular}{ccccc}
        \includegraphics[width=1\linewidth]{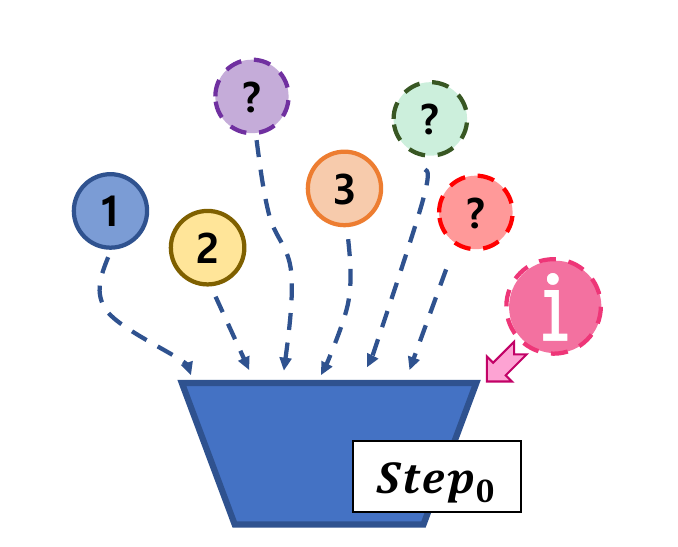} &\includegraphics[width=1\linewidth]{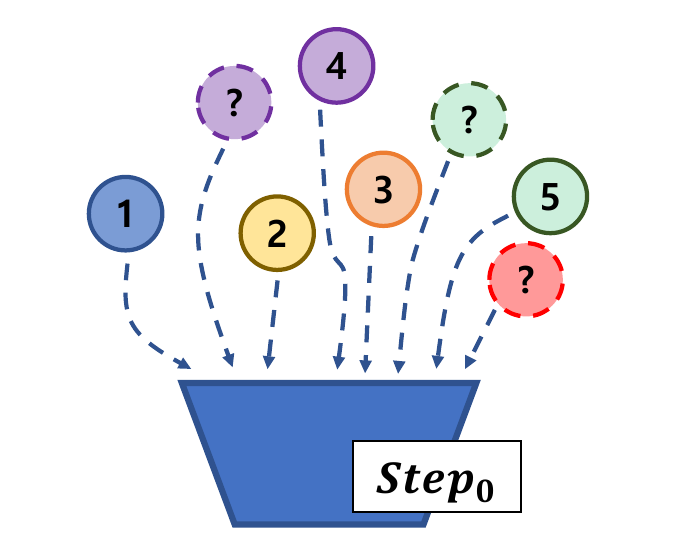}
        &\includegraphics[width=1\linewidth]{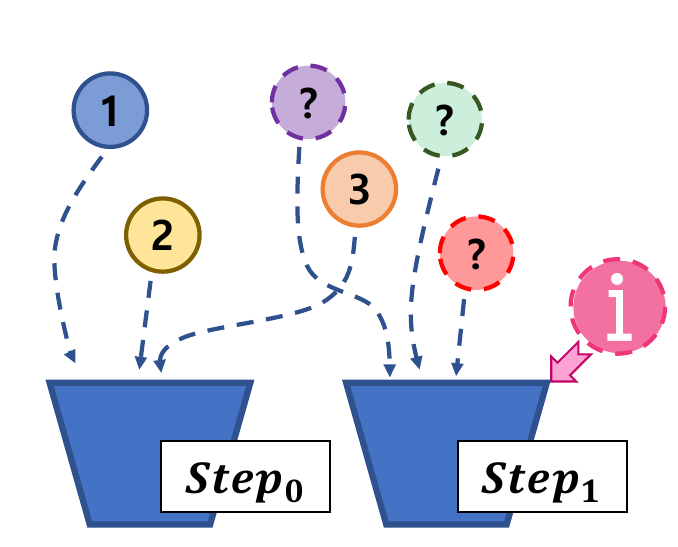}
        &\includegraphics[width=1\linewidth]{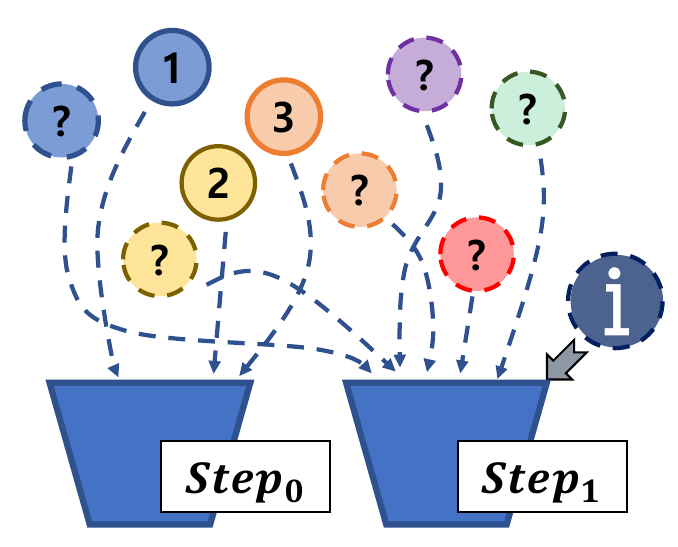}
        &\includegraphics[width=1\linewidth]{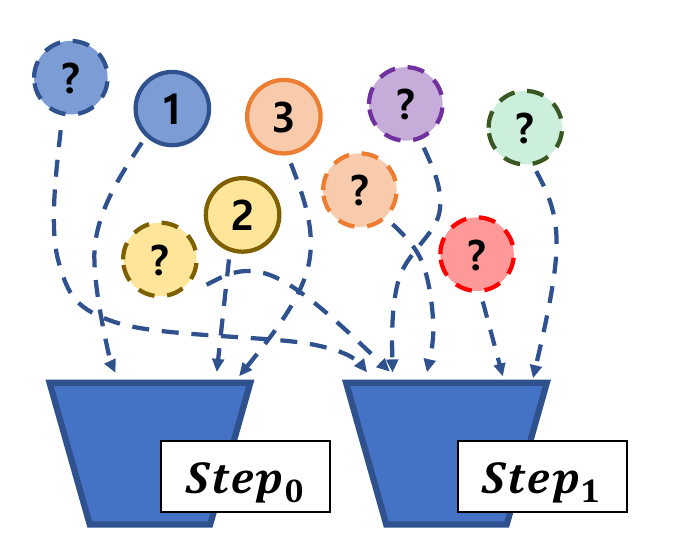}\\
        \fontsize{1.5cm}{1.5cm}\selectfont{(a) NCD} &\fontsize{1.5cm}{1.5cm}\selectfont{(b) GCD} &\fontsize{1.5cm}{1.5cm}\selectfont{(c) class-iNCD} &\fontsize{1.5cm}{1.5cm}\selectfont{(d) CCD-MI} &\fontsize{1.5cm}{1.5cm}\selectfont{(e) CGCD}
    \end{tabular}}
    \vspace{1mm}
    \caption{Comparison of existing and proposed scenarios for novel category discovery. The solid and dash-lined circles indicate labeled and unlabeled samples, respectively, with sample color depicting the class label. The circle with $i$ denotes the number of novel classes (pink) and the proportion of novel class samples in a batch (dark blue). 
    }
    \label{fig:method_scenarios}
    \vspace{-1mm}
\end{figure*}
\begin{table*}[!t]
\small
\begin{center}
\setlength{\tabcolsep}{1pt}
\renewcommand{\arraystretch}{1.0}
\begin{tabular}{p{0.1\linewidth}p{0.25\linewidth}P{0.15\linewidth}P{0.15\linewidth}P{0.15\linewidth}P{0.15\linewidth}}
    \toprule
    \multirow{2}[2]{*}[-1pt]{Task} &\multirow{2}[2]{*}[-1pt]{Method} &\multicolumn{4}{c}{Constraints}\\
    \cmidrule{3-6}
    & &\shortstack{Continual\\learning} &\shortstack{Assumption:\\$\mathcal{Y}_{l}\cap\mathcal{Y}_{u}=\emptyset$} &\shortstack{Number of\\novel classes} &\shortstack{Proportion of\\new class data}\\
    \midrule
    \midrule
    NCD &RankStat~\cite{han2019automatically, han21autonovel}, DRNCD~\cite{zhao21novel} &Not considered &Required &Required &Not required\\
    GCD &GCD~\cite{vaze2022gcd}, XCon~\cite{Fei_2022_BMVC} &Not considered &Not required &Not required &Not required\\
    \midrule
    class-iNCD &FRoST~\cite{roy2022class}, NCDwF~\cite{joseph2022ncdwf} &Considered &Required &Required &Not required\\
    CCD-MI &GM~\cite{zhang2022grow} &Considered &Not required &Not required &Required\\
    \midrule
    CGCD & Ours &Considered &Not required &Not required &Not required\\
    \bottomrule
\end{tabular}
\end{center}
\vspace{-1mm}
\caption{Comparison of existing and proposed novel category discovery settings and the constraints}
\vspace{-3mm}
\label{tab:problem_sota}
\end{table*}

Deep neural networks have achieved remarkable performance in various computer vision tasks. However, current systems are still subject to constraints that are manually supervised and do not consider continual incremental categories. 
For extending to real-world environments,
there are still gaps to catch up by overcoming the constraints and improving their abilities in fundamental tasks.
Specifically, humans still perform better than machines in object cognitive and grouping skills (\eg recognizing new products or clothing on shopping and categorizing undefined moving objects while driving).

Various methods have been proposed to address the limitations of the tasks by considering real-world circumstances, as presented in \cref{fig:method_scenarios} and \cref{tab:problem_sota}.
In detail, Novel Category Discovery (NCD)~\cite{han2019automatically, han21autonovel, zhao21novel} and Generalized Category Discovery (GCD)~\cite{vaze2022gcd, Fei_2022_BMVC} aim to recognize not only pre-trained categories but also discover novel categories using a given dataset. NCD considers a disjoint dataset where labeled and unlabeled novel samples have no intersection with each other. 
In contrast, GCD exploits the joint set with intersected categories, making GCD a more complicated task than NCD. However, these approaches do not consider the class incremental scheme.
Class incremental NCD (class-iNCD)~\cite{roy2022class, joseph2022ncdwf} has been proposed to adopt the incremental categories on NCD task, but they still focus on improving novel category discovery performance using the disjoint set, which is an unrealistic constraint.
To address this issue, Grow and Merge (GM)~\cite{zhang2022grow} proposes a scenario that exploits the joint unlabeled dataset in the incremental novel category discovery task, which is called Continuous Category Discovery Mixed Incremental (CCD-MI).
However, most existing methods require prior knowledge, such as the number of unlabeled classes for NCD and class-iNCD, or the proportion of the novel samples in the batch for CCD-MI. Such prior knowledge requirements are not enough to mimic the real-world, as we lack information about the unlabeled sets.

To overcome these constraints, we propose a novel scenario that better represents real-world circumstances by removing constraints on the available data.
We assume that the given datasets are unlabeled joint sets without providing prior knowledge about the data.
Employing the scenario, we propose a novel unsupervised class incremental learning approach to simultaneously address the problems of discovering incremental novel categories and alleviating catastrophic forgetting.
In addition, we focus on recognizing fine-grained objects, which is a more realistic use case for various applications in real-world applications. 

The proposed method exploits a deep metric learning scheme, proxy anchor (PA)~\cite{Kim_2020_CVPR} that presents fast and reliable convergence and robustness against noise samples, and also considers the relations between samples. Then, we divide the unlabeled data into old and novel categories using PAs, which inherit discriminative features of old categories. The cosine similarity is measured between the PAs and the samples, and then initially separated datasets are acquired on a criterion.
For further splitting, we adopt a noisy label learning scheme, and then assign the predictions of the previous model and the clustering results by a non-parametric approach to old and novel categorized samples, respectively.
To mitigate the forgetting,
we use a PA-based exemplar, which inherits more representative features.
In the experimental results, we demonstrate that the proposed method outperforms the existing state-of-the-art in discovering novel categories and forgetting alleviation on various fine-grained datasets.
Specifically, the proposed method does not require any prior knowledge and considers continual learning on unlabeled joint datasets, making it a more realistic and practical solution for real-world scenarios.

The main contributions of the proposed method can be summarized as follows.
\begin{itemize}
    \item We introduce a novel scenario, called Continuous Generalized novel Category Discovery (CGCD), which is well-suited to tackle the challenges of discovering novel categories in real-world scenarios by removing the constraint that unlabeled data belong to only novel categories.
    \item We propose a novel unsupervised learning approach for incremental novel category discovery that does not require prior knowledge of the number of novel categories or the proportion of new class data.
    \item We present a noisy label learning approach and deep metric learning to split unlabeled data into old and novel categories, and also show mitigation of catastrophic forgetting using a deep metric-based exemplar.
    \item The proposed method outperforms the state-of-the-art methods in novel category discovery and forgetting mitigation on various fine-grained datasets.
\end{itemize}
\begin{figure*}[t]
    \centering
    \resizebox{1.\linewidth}{!}{
    \setlength{\tabcolsep}{1pt}
    \begin{tabular}{c}
        \includegraphics[width=1\linewidth]{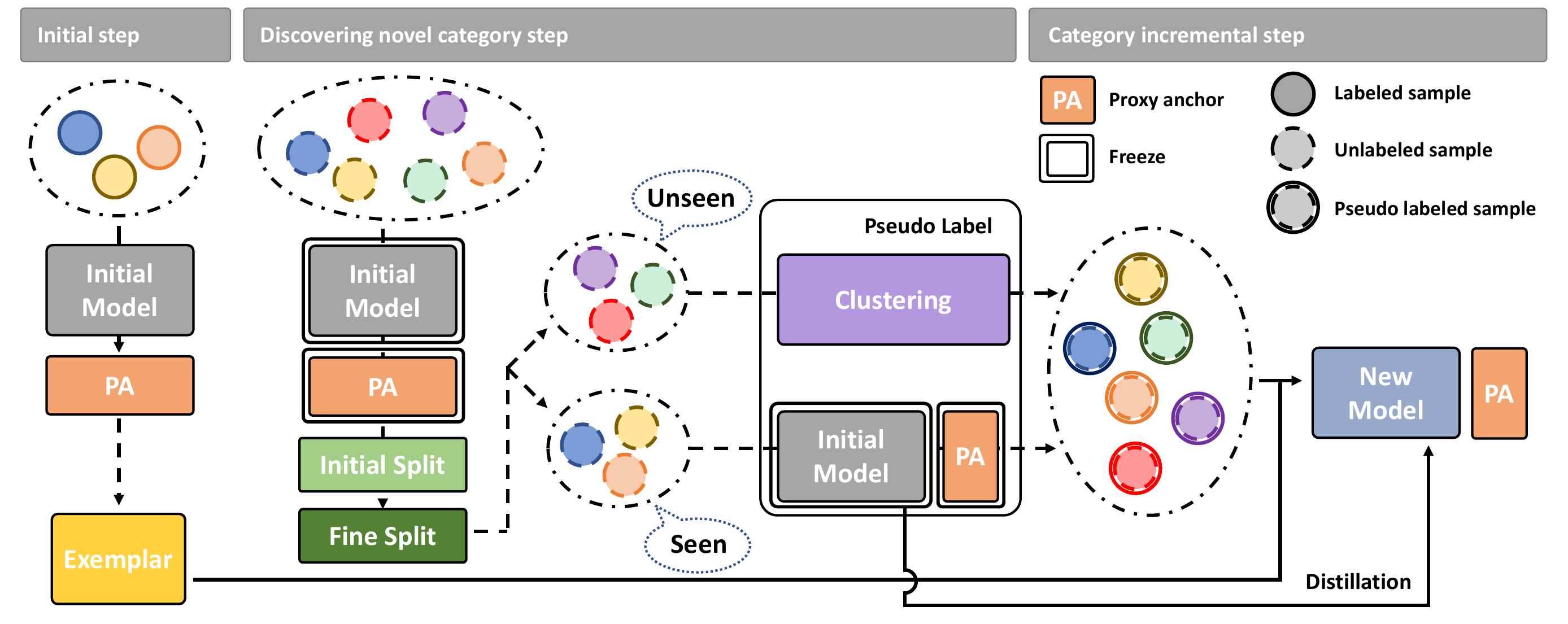}
    \end{tabular}}
    \caption{Overview of the proposed CGCD framework. The framework comprises three steps. The first step is that the network model is fine-tuned on the target dataset using the labeled dataset. In the second step, the discovery of novel categories is performed on the joint and unlabeled dataset, which is split into unseen and seen sets using the initial and fine split methods. 
    Pseudo-labels are assigned to the unlabeled dataset using the previous model's predictions (seen case) and non-parametric clustering results (unseen case). In the last step, a new model is trained on the fine split dataset, which incorporates new proxy anchors based on clustering results. 
    }
    \vspace{-4mm}
    \label{fig:method_overview}
\end{figure*}

\vspace{-3mm}
\section{Problem Definition}\label{sec:problem}
\subsection{Continuous Generalized Category Discovery}\label{sec:problem_setting}
As presented in \cref{fig:method_scenarios} and \cref{tab:problem_sota}, various environmental schemes have been proposed to mimic real-world circumstances. Notable approaches include NCD \cite{han2019automatically, han21autonovel, zhao21novel}, GCD \cite{vaze2022gcd}, class-iNCD \cite{roy2022class, joseph2022ncdwf}, and CCD \cite{zhang2022grow}. NCD considers disjoint datasets between labeled and unlabeled sets (\ie$\mathcal{Y}_l\cap\mathcal{Y}_u=\emptyset$) and requires prior knowledge of the number of unlabeled categories $\lvert{\mathcal{Y}_u}\rvert$. In contrast, GCD exploits the joint set (\ie$\mathcal{Y}_l\cap\mathcal{Y}_u\neq\emptyset$). Although GCD is a more challenging task than NCD, it still does not consider continuous incremental category discovery.

Class-iNCD is an extension of NCD to the continual learning scheme. However, the method trains on the disjoint dataset under class incremental stages and has a requirement for $\lvert{\mathcal{Y}_u}\rvert$. CCD also trains on discovering novel classes under the continual learning scheme with the joint dataset. Although CCD does not require $\lvert{\mathcal{Y}_u}\rvert$,
it still needs the proportion of the novel class data in a batch as prior knowledge, which is used to filter out the data.

To address these limitations, we propose a more challenging problem, named Continuous Generalized novel Category Discovery (CGCD), that is closer to real-world circumstances.
In CGCD, we exploit the unlabeled joint dataset in incremental steps without providing any prior knowledge and aim to discover novel classes. This formulation is more representative of real-world scenarios, where we are not aware of the number of unlabeled categories and the characteristics of the dataset.

\subsection{Setting of Continuous Generalized Category Discovery}
\label{sec:problem_dataset}
Fine-grained datasets consist of similar objects, such as canines~\cite{khosla2011novel}, indoor scenes~\cite{quattoni2009recognizing}, vehicles~\cite{maji2013fine}, and birds~\cite{wah2011caltech}, in constrained circumstances. Compared to coarse datasets such as CIFAR~\cite{krizhevsky2009learning} and ImageNet~\cite{krizhevsky2012imagenet}, fine-grained datasets are closer to real-world scenarios. Therefore, we focus on training and discovering novel classes with fine-grained datasets to mimic real-world circumstances better.

CGCD employs the joint unlabeled dataset in incremental steps, and
\cref{tab:problem_dataset} describes the dataset partitioning used in the one-time incremental category discovery.
First, the set of classes is partitioned into old classes and new classes at a certain rate, for example, $8\colon{2}$. The initial step utilizes labeled dataset $\mathcal{D}^0$ consisting of old classes only. Then the following incremental step utilizes unlabeled dataset $\mathcal{D}^1$ consisting of both old classes and new classes, which reflect more realistic and challenging real-world scenarios. The key element of the proposed method is to decide whether the unlabeled data point belongs to the old classes (seen) or new classes (unseen).
The samples belonging to the old classes are assigned to labeled dataset $\mathcal{D}^0$ and unlabeled dataset $\mathcal{D}^1$, and
the rest of all the new class samples are assigned to $\mathcal{D}^1$. Here the choice of $8\colon{2}$ is an arbitrary example, and it is important to note that this ratio is just for data generation purposes and is not revealed to the learning methods.
\begin{table}[!t]
\small
\begin{center}
\setlength{\tabcolsep}{1pt}
\renewcommand{\arraystretch}{1.0}
\begin{tabular}{p{0.23\columnwidth}P{0.09\columnwidth}P{0.01\columnwidth}P{0.2\columnwidth}P{0.01\columnwidth}P{0.2\columnwidth}P{0.18\columnwidth}}
    \toprule
    \multirow{2}[2]{*}[-1pt]{Dataset} &\multirow{2}[2]{*}[-1pt]{\shortstack{All\\classes}} & &Initial step & &\multicolumn{2}{c}{Category incremental step}\\    
    \cmidrule{4-4}\cmidrule{6-7}
    & & &Old class & &Old class &New class\\
    \midrule
    \midrule
    CUB-200 &200 & &160 (0.8) & &160 (0.2) &40 (1.0)\\
    MIT67 &67 & &53 (0.8) & &53 (0.2) &14 (1.0)\\
    Stanford Dogs &120 & &96 (0.8) & &96 (0.2) &24 (1.0)\\
    FGVC Aircraft &100 & &80 (0.8) & &80 (0.2) &20 (1.0)\\    
    \bottomrule
\end{tabular}
\end{center}
\caption{Dataset configurations for one-time incremental category scenarios. The number of classes and the proportion of data from each class in parentheses are presented. 
Note that the ratios in parentheses are hidden information that is not revealed to the learning methods.}
\vspace{-5mm}
\label{tab:problem_dataset}
\end{table}

\section{Method}\label{sec:method}
As described in~\cref{fig:method_overview}, our proposed method consists of three steps: the initial step, the novel category discovery step, and the category incremental step. In the initial step, we fine-tune a pre-trained model on the labeled dataset $\mathcal{D}^0=\{(x, y)\in\mathcal{X}_l\times \mathcal{Y}_l\}$, and obtain the embedding vector $z$ using the model $f(\cdot)$, denoted as $z=f^0(x)$. We then use these vectors to train PAs~\cite{Kim_2020_CVPR} of each category, represented as ${p}=g^0(z)$, and also construct well-representative exemplars.
In the following novel category discovery step, the given unlabeled joint datasets are denoted as $\mathcal{D}^1=\{x \vert x\in\mathcal{X}_u\}$.
We first separate them into old and novel categories through the initial and the fine splits. 
Since the separated sets are unlabeled, we pseudo-label for old and novel classes using the previous model prediction and non-parametric clustering results, respectively.

In the category incremental step, the acquired set is trained to improve the performance of discovering novel categories. To avoid catastrophic forgetting, we exploit generated features by the exemplar and feature distillation between earlier and new models. The proposed model does not require any prior knowledge, such as $\lvert\mathcal{Y}_u\rvert$ and the ratio of novel class samples in a batch.
We evaluate the performance of the proposed method using the validation dataset, which includes all categories.

\subsection{Initial Step: Fine Tune}\label{sec:method_proxy}
Existing NCD methods do not account for noisy categories, such as those categorized from old to novel or from novel to old, which can impair novel discovery performance and accumulate errors in the continuous procedure. To address these limitations, in this work, we propose a novel approach that leverages the benefits of PA to complement and improve the existing approaches. PA is a metric learning method that combines proxy- and pair-based methods to achieve rapid and reliable convergence, and robustness against noisy samples, and considers relations between data to extract meaningful semantic information.

Following the method, the embedding vector $z$ from the initial model $f^0$ is trained to map to each proxy anchor $p=g^0(z)$.
Let the set of all proxy anchors as $P^0$ in the labeled data $\mathcal{D}^0$.
In this manner, the number of proxy anchors of $\mathcal{D}^0$ is the number of classes of the labeled set (\ie$\lvert{P^0}\rvert=\lvert\mathcal{Y}_l\rvert$) in the initial step. We train the model and proxy anchors using the following loss function defined in~\cite{Kim_2020_CVPR}:
\begin{equation}
\resizebox{0.9\columnwidth}{!}{
    $
    \begin{aligned}
        \mathcal{L}^0_{pa}(Z^0)&=\frac{1}{\lvert{P^{0^+}}\rvert}\sum_{p\in{P^{0^+}}}\log\bigg(1+\sum_{z\in{Z^{0^+}_p}}{e^{-\alpha(s(z, p)-\delta)}}\bigg)\\        
        &+\frac{1}{\lvert{P^{0}}\rvert}\sum_{p\in{P^0}}\log\bigg(1+\sum_{z\in{Z^{0^-}_p}}{e^{\alpha(s(z, p)+\delta)}}\bigg)
    \end{aligned}
    $
}
\label{eq:method_proxy}
\end{equation}

\noindent where $\delta>0$ is a margin and $\alpha>0$ is a scaling factor.
The function $s(\cdot{,}\cdot)$ indicates the cosine similarity score.
$P^{0^+}$ represents same class PAs(\eg{positive}) in the batch. Each proxy $p$ divides the set of embedding vector $Z^0$ as $Z^{0^+}_p$ and $Z^{0^-}_p=Z^0-Z^{0^+}_p$. $Z^{0^+}_p$ denotes the same class embedding points with the proxy anchor $p$.
The first term aims to pull $p$ and its dissimilar but hard positive data together, while the last term is to push $p$ and its similar but hard negatives apart.

\subsection{Discovering Novel Categories Step}\label{sec:method_nd}
\noindent\textbf{Separation:} 
In this procedure, we aim to split the given joint dataset $\mathcal{D}^1$ into the novel and old categories without any prior knowledge. We conduct this task in two stages: initial split and fine split.
In the initial split, we compute the cosine similarity between $p$ and each embedding vector $z_i\in Z^1$, where $z_i=f^0(x_i)$ and $x_i\in D^1$. Because the set of proxy anchors $P^0$ represents the old categories, we classify a sample to the old class if the maximum similarity score of $z_i$ is larger than a threshold $\epsilon$. We set $\epsilon=0$ since it is the median of the score ranges. 
The initial split is defined as:
\begin{figure}[t]
    \centering
    \resizebox{1.\columnwidth}{!}{
    \setlength{\tabcolsep}{1pt}
    \begin{tabular}{cc}
        \includegraphics[width=1\columnwidth]{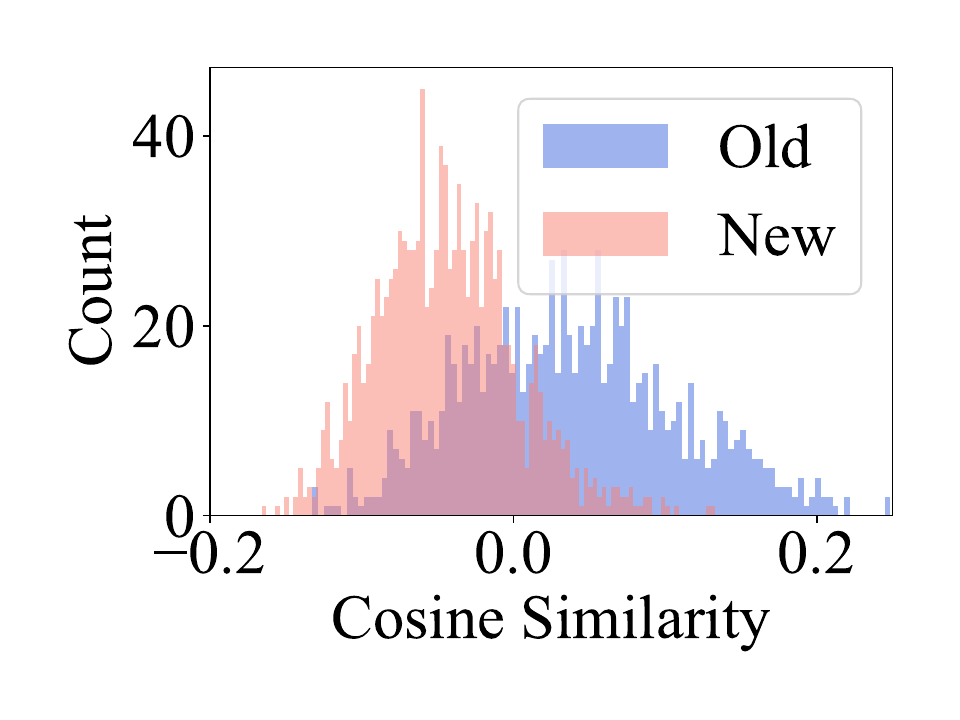} &\includegraphics[width=1\columnwidth]{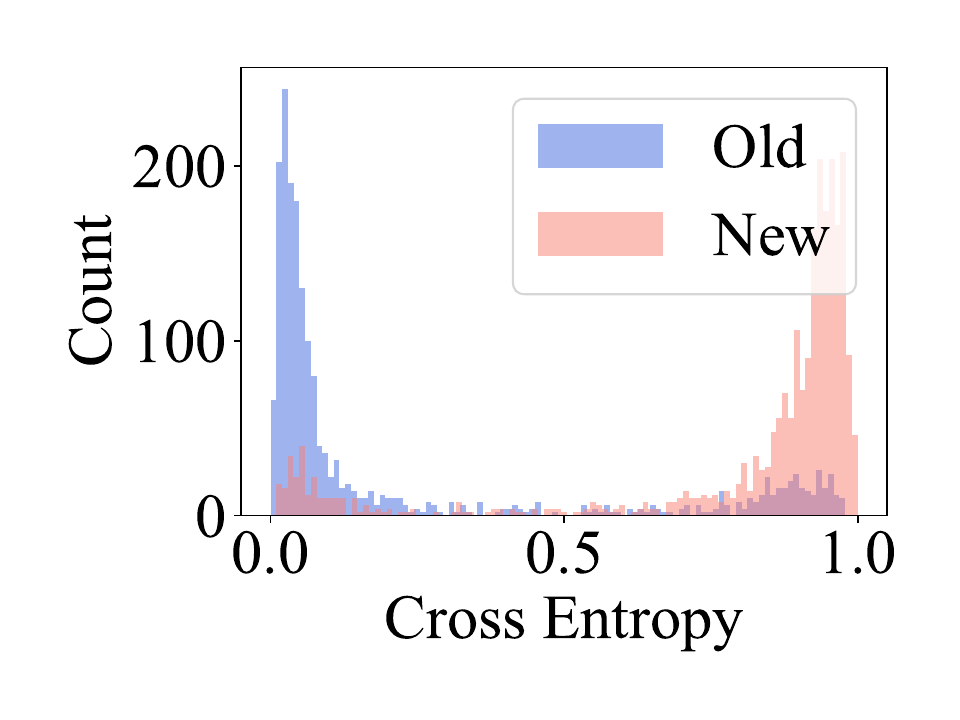}\\
        \fontsize{0.6cm}{0.6cm}\selectfont{(a) Initial Split} &\fontsize{0.6cm}{0.6cm}\selectfont{(b) Fine Split}        
    \end{tabular}}
    \caption{Separation results on joint unlabeled datasets}
    \vspace{-5mm}
    \label{fig:ablation_nl}
\end{figure}
\begin{equation}
\begin{aligned}
    \Tilde{y}_{i}=
    \begin{cases}
      0, &\text{if $\underset{p\in{P^0}}{\max}(s(z_i, p))\geq{\epsilon}=0$}\\
      1, &\text{otherwise}\\      
    \end{cases} 
\end{aligned}
\label{eq:method_init_split}
\end{equation}

To acquire a cleaner novel and old dataset, we propose a noisy labeling scheme, fine split, which involves an iterative training of a simple multilayer perceptron (MLP) based classifier $m(\cdot)$ on the binarized dataset. 
The initial split results in noisy and inaccurate separation, as shown in \cref{fig:ablation_nl} (a). 
Among them, only the data on both ends of the spectrum are assumed the clean and utilized to train the classifier.
The loss of split network $m(\cdot)$ is defined as follows:
\begin{equation}
\resizebox{0.9\columnwidth}{!}{
$
\begin{aligned}
    \mathcal{L}_{sp}=-\mathbb{E}_{{z_c}\in{Z^1_c}}[\Tilde{y}_{c}\log(m(z_c))+(1-\Tilde{y}_{c})\log(1-m(z_c))]\\
\end{aligned}
$
}
\label{eq:method_bce}
\end{equation}

where $z_c$ denotes clean embedding vectors, and $Z^1_c$ represents the set of clean vectors. $\Tilde{y}_{c}$ indicates the pseudo label of the clean data.
After the warm-up training, the classifier is trained with re-assigned pseudo labels and the more cleaned data, which is divided with the Gaussian mixture model (GMM).
Consequently, the \cref{fig:ablation_nl} (b) shows the cleaner separated results.

\noindent\textbf{Pseudo-labeling:} 
After the separation, both the old $\mathcal{D}_{old}^1$ and the novel categories $\mathcal{D}_{new}^1$ are still unlabeled. Thus, we use pseudo-labels to assign labels to each sample. For $\mathcal{D}_{old}^1$, we use the predictions of the previous model and proxy anchors to assign pseudo-labels. In contrast, we use a non-parametric clustering approach named Affinity propagation~\cite{frey2007clustering} to assign pseudo-labels to $\mathcal{D}_{new}^1$. In this manner, our proposed approach does not require any prior knowledge. Finally, from the clustering results, we obtain an estimate of the number of novel categories, denoted as $\lvert\mathcal{\hat{Y}}_n\rvert$.

\subsection{Category Incremental Step}\label{sec:method_cil}
\noindent\textbf{Training modified model and PAs:} 
To improve the performance of discovering novel categories, we modify the model since the previous model has PAs only for $\lvert\mathcal{Y}_l\rvert$ classes and cannot categorize novel classes. We add new $p$ for $\lvert\mathcal{\hat{Y}}_n\rvert$ classes, increasing the total number of PAs like $\lvert{P^1}\rvert=\lvert\mathcal{Y}_l\rvert+\lvert\mathcal{\hat{Y}}_n\rvert$. The loss function to train the modified model $f^1$ and PA $p=g^1(z)$ is reformulated as follows:
\begin{equation}
\resizebox{0.9\columnwidth}{!}{
    $
    \begin{aligned}
        \mathcal{L}^{1}_{pa}(Z^1)&=\frac{1}{\lvert{P^{1^+}}\rvert}\sum_{p\in{P^{1^+}}}\log\bigg(1+\sum_{z\in{Z^{1^+}_p}}{e^{-\alpha(s(z, p)-\delta)}}\bigg)\\        
        &+\frac{1}{\lvert{P^1}\rvert}\sum_{p\in{P^1}}\log\bigg(1+\sum_{z\in{Z^{1^-}_p}}{e^{\alpha(s(z, p)+\delta)}}\bigg)
    \end{aligned}
    $
}
\label{eq:method_proxy_increased}
\end{equation}

\noindent\textbf{Avoiding forgetting:}
In the continual learning scheme, it is essential to alleviate catastrophic forgetting. We adopt feature replay, which leverages the PA information belonging to old categories. Each well-trained $p$ inherits the representation power for each category. We employ each $p$ to generate features by following the Gaussian distribution $\mathcal{N}(p^0, \sigma^2), p^0\in P^0$. The number of generated features is determined based on data balancing, for example, the number of newly categorized samples in a batch. The generated features are concatenated into a batch, and the model and PAs are trained using the following loss function:
\begin{equation}
\begin{aligned}
    \mathcal{L}^{1}_{ex}(\tilde{Z})=\mathcal{L}^{1}_{pa}(\tilde{Z}), \quad\quad\tilde{Z}=\{\tilde{z}\sim\mathcal{N}(p^0,\sigma^2)\}\\    
\end{aligned}
\label{eq:method_exempler}
\end{equation}

Also, we utilize the distillation of the extracted embedding vectors from the present $f^1$ and the previous model $f^0$. The distillation loss $\mathcal{L}_{kd}$ is described as follows:
\begin{equation}
\begin{aligned}
    \mathcal{L}^1_{kd}(z_{o}) &= -\mathbb{E}_{z_{o}\in Z^1_{old}}\|{z^0_o-z_{o}\|_2}\\
    &=-\mathbb{E}_{x_{o}\in D^1_{old}}\|{f^0(x_{o})-f^1(x_{o})}\|_2
\end{aligned}
\label{eq:method_kd}
\end{equation}
where $z^0$ represents the embedding vector for fixed previous feature network $f^0$. $Z^1_{old}$ denotes seen data from $Z^1=\{Z^1_{old}\cup Z^1_{new}\}$. 

In conclusion, the loss consists of three different losses in the continuous category discovery step. One loss is for training the PAs and the model on $\mathcal{D}^1$, the others are to avoid forgetting by using generated features and knowledge distillation. $\mathcal{L}^1$ is described as follows:
\begin{equation}
\begin{aligned}
    \mathcal{L}^1 = \mathcal{L}^1_{pa}(Z^1)+\mathcal{L}^1_{ex}(\tilde{Z})+\mathcal{L}^1_{kd}(z_{o})
\end{aligned}
\label{eq:method_loss_step}
\end{equation}

\section{Experimental Results}\label{sec:experiment}
\subsection{Implementation Details}\label{sec:experiment_detail}
We utilized the widely used augmentation techniques, including random crop after padding and random horizontal flip. All the experiments were trained for $60$ epochs using AdamW optimizer with weight decay set to 0.0001. The initial learning rate was set to 0.0001 for the model $f(\cdot)$, while for the PAs, it was set to 0.01. The learning rate was decayed by a factor of 0.5 every five epochs. We used the threshold $\epsilon$ only once for the initial split to divide the set into old and novel categories, and we set it to 0 for all the datasets and networks. For fine split, we used an MLP-based network architecture that consists of two dense layers with a batch normalization layer.
The model was trained for three epochs using AdamW with a learning rate of 0.0001. The hyperparameters for PAs, $\alpha$ and $\delta$, were set to 32 and 0.1, respectively.

For fair comparisons of various methods, such as NCD, class-iNCD, and CCD, we follow the hyperparameters and the network architectures of the original implementations, referring to the papers for details. All the reported performances are average results over three runs.

\subsection{Evaluation Metrics}\label{sec:problem_metric}
We evaluate the methods using metrics based on the cluster accuracy measurement, called Hungarian assignment algorithm~\cite{Kuhn1955Hungarian}.
The evaluation metric is defined as follows:
\begin{equation}
\begin{aligned}
    \mathcal{M}^t=\frac{1}{\lvert{\mathcal{D}}\rvert}\sum_{i=1}^{{\lvert{\mathcal{D}}\rvert}}\mathbb{I}(y_{i}=h^{\ast}(y^{\ast}_{i}))
\end{aligned}
\label{eq:metric_hungarian}
\end{equation}
where ${\lvert{\mathcal{D}}\rvert}$ is the size of the validation dataset $\mathcal{D}$ and $h^\ast$ is the optimal assignment. So, $\mathcal{M}^t$ measures the cluster accuracy at step $t$ on the $\mathcal{D}$. In this manner, $\mathcal{M}_{all}$ and $\mathcal{M}_{o}$ indicate the cluster accuracy metrics of the whole and old categories using $\mathcal{M}^t$, respectively.
Furthermore, we employ two more metrics that, $\mathcal{M}_f$ and $\mathcal{M}_d$, which are proposed on GM~\cite{zhang2022grow} and described as follows:
\begin{equation}
\begin{aligned}
    \mathcal{M}_f = \max_{t}\{\mathcal{M}^{0}_{o}-\mathcal{M}^{t}_{o}\},
\end{aligned}
\label{eq:metric_mf}
\end{equation}
\begin{equation}
\begin{aligned}
    \mathcal{M}_d=\frac{1}{\lvert{T}\rvert}\sum_{i=T}\mathcal{M}^{i}_{n}.
\end{aligned}
\label{eq:metric_md}
\end{equation}
\begin{table*}[t] 
\small
\begin{center}
\setlength{\tabcolsep}{1pt}
\renewcommand{\arraystretch}{1.0}
\begin{tabular}{p{0.1\linewidth}P{0.05\linewidth}P{0.05\linewidth}P{0.05\linewidth}P{0.05\linewidth}P{0.01\linewidth}P{0.05\linewidth}P{0.05\linewidth}P{0.05\linewidth}P{0.05\linewidth}P{0.01\linewidth}P{0.05\linewidth}P{0.05\linewidth}P{0.05\linewidth}P{0.05\linewidth}P{0.01\linewidth}P{0.05\linewidth}P{0.05\linewidth}P{0.05\linewidth}P{0.05\linewidth}}
    \toprule
    \multirow{2}[2]{*}[-1pt]{~Method} &\multicolumn{4}{c}{CUB-200} & &\multicolumn{4}{c}{MIT67} & &\multicolumn{4}{c}{Stanford Dogs} & &\multicolumn{4}{c}{FGVC aircraft}\\
    \cmidrule{2-5}\cmidrule{7-10}\cmidrule{12-15}\cmidrule{17-20}
    &{$\mathcal{M}_{all}\uparrow$} &{$\mathcal{M}_{o}\uparrow$} &{$\mathcal{M}_{f}\downarrow$} &{$\mathcal{M}_{d}\uparrow$} & &{$\mathcal{M}_{all}\uparrow$} &{$\mathcal{M}_{o}\uparrow$} &{$\mathcal{M}_{f}\downarrow$} &{$\mathcal{M}_{d}\uparrow$} & &{$\mathcal{M}_{all}\uparrow$} &{$\mathcal{M}_{o}\uparrow$} &{$\mathcal{M}_{f}\downarrow$} &{$\mathcal{M}_{d}\uparrow$} && {$\mathcal{M}_{all}\uparrow$} &{$\mathcal{M}_{o}\uparrow$} &{$\mathcal{M}_{f}\downarrow$} &{$\mathcal{M}_{d}\uparrow$}\\
    \midrule
    \midrule
    ~Supervised &61.69 &45.83 &23.32 &16.63 & &55.56 &40.90 &19.52 &18.34 & &64.26 &47.57 &25.43 &17.01 & &64.62 &48.17 &26.24 &18.12\\    
    ~DRNCD~\cite{zhao21novel} &9.80 &10.47 &58.51 &34.24 & &26.99 &27.67 &49.07 &38.91 & &16.39 &10.34 &65.83 &63.36 & &18.73 &19.50 &57.05 &45.63\\
    ~FRoST~\cite{roy2022class} &18.19 &17.34 &12.18 &17.20 & &23.21 &23.55 &15.96 &24.74 & &22.62 &22.23 &14.29 &26.06 & &32.61 &33.53 &15.19 &27.69\\
    \midrule
    ~GM~\cite{zhang2022grow} &6.43 &6.57 &39.82 &5.92 & &16.52 &16.77 &42.26 &15.50 & &5.99 &5.98 &50.36 &5.98 & &12.00 &11.63 &53.35 &13.46 \\
    ~Ours &\textbf{54.75} &\textbf{58.80} &\textbf{15.47} &\textbf{40.90} & &\textbf{54.45} &\textbf{64.23} &\textbf{10.64} &\textbf{18.58} & &\textbf{66.25} &\textbf{76.15} &\textbf{6.77} &\textbf{30.04} & &\textbf{37.28} &\textbf{41.60} &\textbf{14.66} &\textbf{20.04}\\    
    \bottomrule
\end{tabular}
\end{center}
\vspace*{-1mm}
\caption{Comparison results under continuous generalized categorized discovery scenario. The results present the mean over three runs.}
\vspace*{-3mm}
\label{tab:experiment_sota}
\end{table*}
\begin{figure*}[t]
    \centering
    \small
    \resizebox{1\linewidth}{!}{
    \setlength{\tabcolsep}{1.pt}
    \begin{tabular}{ccccccc}
        {\includegraphics[width=1.\linewidth]{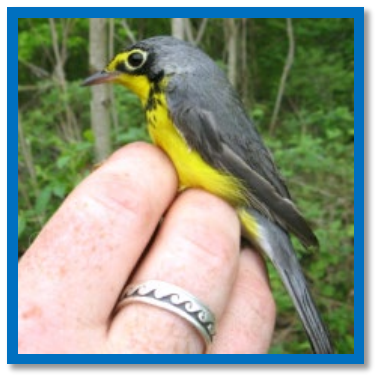}} &{\includegraphics[width=1.\linewidth]{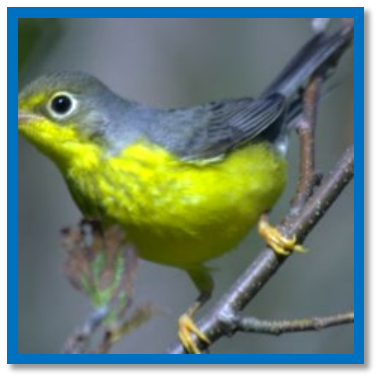}} &{\includegraphics[width=1.\linewidth]{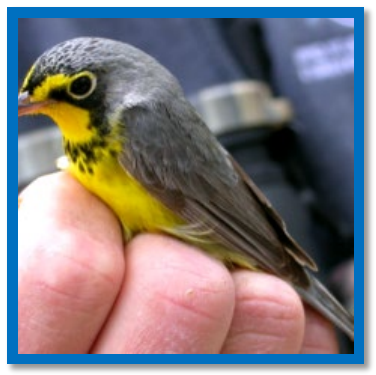}} &{\includegraphics[width=1.\linewidth]{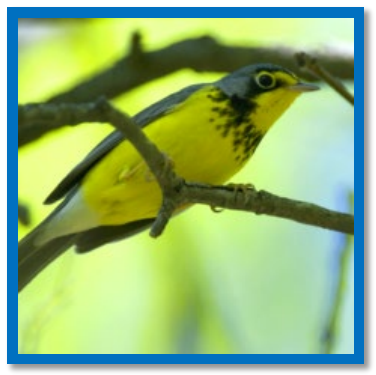}} &{\includegraphics[width=1.\linewidth]{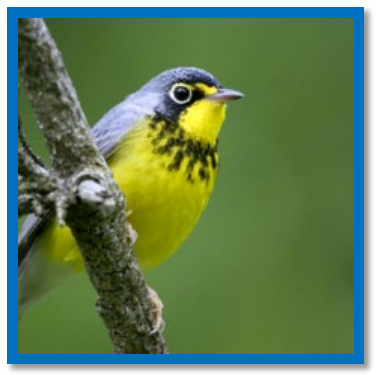}} &{\includegraphics[width=1.\linewidth]{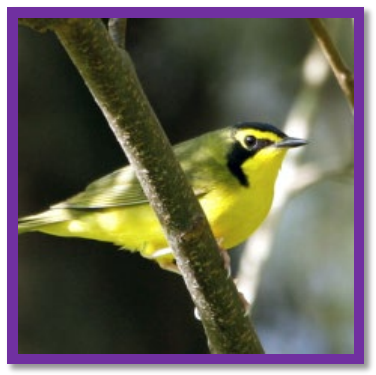}} &{\includegraphics[width=1.\linewidth]{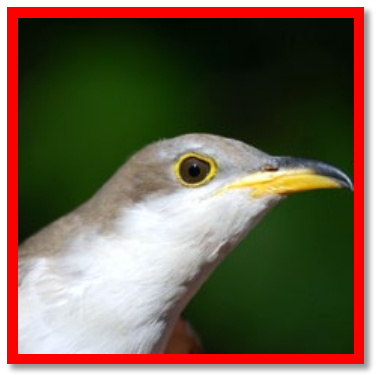}}\\
        {\includegraphics[width=1.\linewidth]{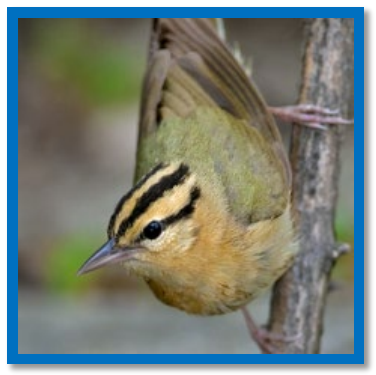}} &{\includegraphics[width=1.\linewidth]{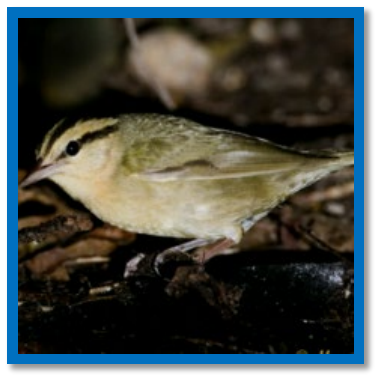}} &{\includegraphics[width=1.\linewidth]{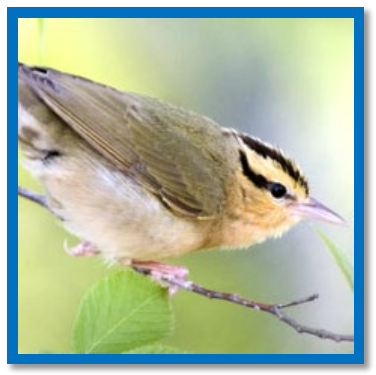}} &{\includegraphics[width=1.\linewidth]{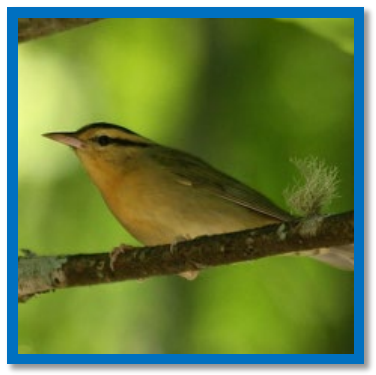}} &{\includegraphics[width=1.\linewidth]{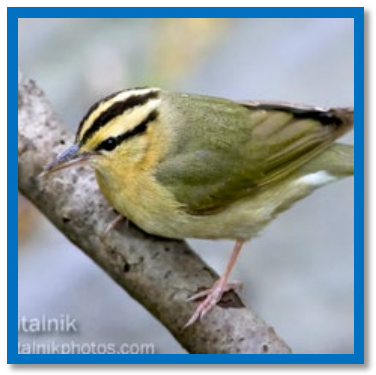}} &         {\includegraphics[width=1.\linewidth]{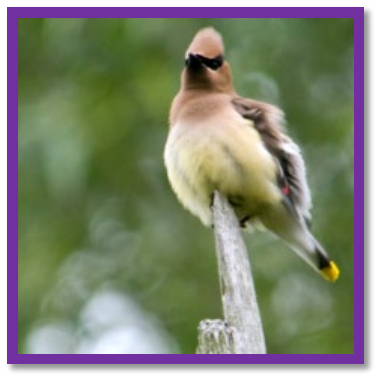}} & {\includegraphics[width=1.\linewidth]{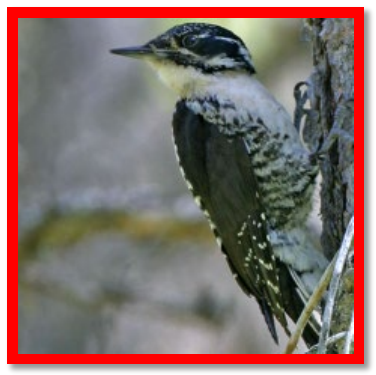}}\\
        {\includegraphics[width=1.\linewidth]{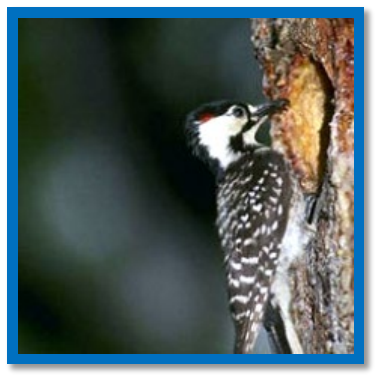}} &{\includegraphics[width=1.\linewidth]{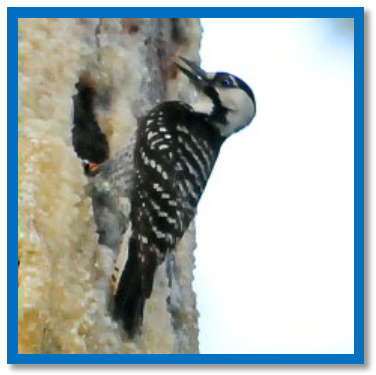}} &{\includegraphics[width=1.\linewidth]{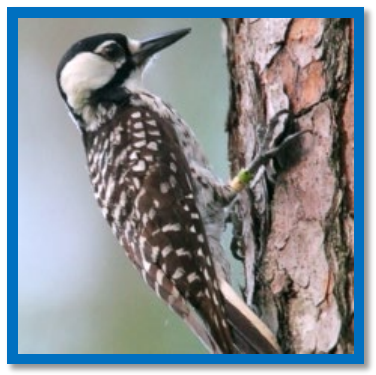}} &{\includegraphics[width=1.\linewidth]{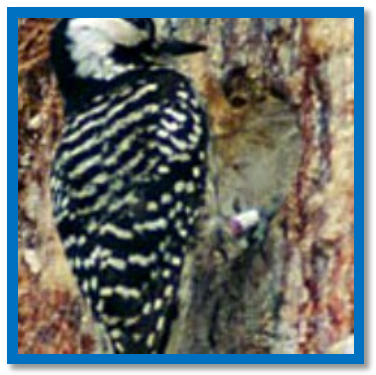}} &{\includegraphics[width=1.\linewidth]{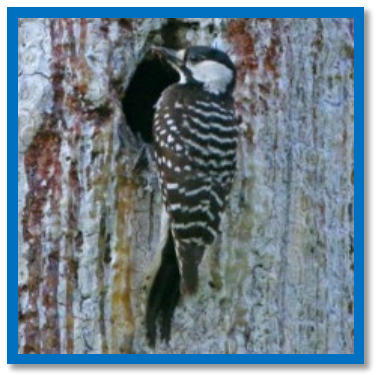}} &         {\includegraphics[width=1.\linewidth]{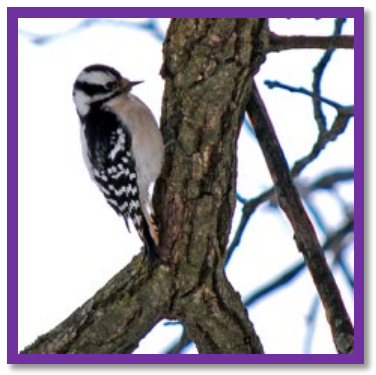}} & {\includegraphics[width=1.\linewidth]{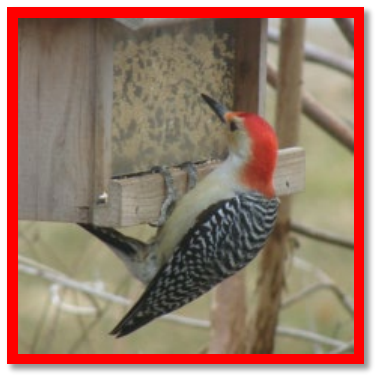}}\\
    \end{tabular}}
    \caption{
    Qualitative evaluation results of the proposed method using the CUB-200 dataset on ResNet-18. The first five columns with blue boxes denote well-clustered examples. The last two columns represent failed prediction results, including example images with purple boxes denoting hard negatives and those with red boxes indicating incorrect categorization.}
    \vspace*{-3mm}    
    \label{fig:experiment_quality}
\end{figure*}

\noindent where $\mathcal{M}^{0}_o$ and $\mathcal{M}^{t}_o$ are the old class cluster accuracy at the initial step and category incremental $t$ step, respectively. So $\mathcal{M}_f$ measures the maximum forgetting values for old categories in the entire step and should be sufficiently low; if not, the method is not valuable in practical applications.
In~\cref{eq:metric_md}, $\lvert{T}\rvert$ is the number of increased steps, and $\mathcal{M}_d$ evaluates the averaged performance of novel category discovery in unlabeled joint datasets in each step. It means the higher method, the more appropriate method in real-world applications.

\subsection{Comparison with State-of-the-arts Methods}\label{sec:experiment_sota}
We conducted a series of experiments to compare the cluster accuracy performance of our proposed method with state-of-the-art approaches, as presented in \cref{tab:problem_sota}. In the experiments, we excluded the GCD task, including XCon, as it focuses on discovering novel categories blended with labeled and unlabeled datasets and evaluates unknown discovery performance on only train datasets, not validation datasets, in their papers. Therefore, we compared our proposed method with other approaches, including Dual rank NCD (DRNCD)~\cite{zhao21novel}, FRoST~\cite{roy2022class}, and GM, which are the representative methods that record the state-of-the-art results of each task.

We first evaluated the one-step incremental category setting and reported the comparison results in \cref{tab:experiment_sota}. The supervised method is the setting of a supervised continual learning manner, literally. We observed catastrophic forgetting in supervised learning. DRNCD is one of the NCD approaches and recorded the outstanding performances of $\mathcal{M}_d$ on MIT67, Stanford Dogs, and FGVC aircraft. However, the method requires prior knowledge of the number of novel categories and does not identify specific classes, but only knows whether the samples are included in the novel category or not. Thus, the results are not regarded as outperforming. Instead, the method shows the highest results of $\mathcal{M}_f$, which means that the method focuses on learning novel categories without considering the prevention of forgetting previous categorical knowledge. In this regard, NCD is not sufficient to extend novel class incremental learning schemes.
FRoST showed competitive results of discovering novel classes and decreased forgetting results $\mathcal{M}_f$ on all datasets compared to DRNCD.
Nevertheless, considering the required knowledge, the other metrics results, $\mathcal{M}_{all}$ and $\mathcal{M}_o$ were not competitive. GM proposed the CCD setting, which is the most similar to CGCD, in the perspective of the unknown number of novel categories and discovering novel and old categories on new unlabeled datasets. However, the method requires a crucial parameter, which is the ratio of novel category samples on the new dataset. Without considering the ratio, GM recorded the lowest results of $\mathcal{M}_{all}$, $\mathcal{M}_o$, and $\mathcal{M}_d$. The proposed method showed outstanding performance on the various datasets without requiring any prior information about new incoming unlabeled datasets. Our method recorded the second-best $\mathcal{M}_f$ on the CUB-200 dataset and the best $\mathcal{M}_f$ on the other datasets, such as MIT67, Dogs, and FGVC aircraft, by the effects of PA-based exemplar. On $\mathcal{M}_d$, the method was also competitive and compared to the GM, which has the most similar setting to our method.

To evaluate the proposed method qualitatively, we clustered the evaluation dataset using the CUB-200 dataset. 
In~\cref{fig:experiment_quality}, our method well-discovered novel categories and clustered them correctly. Each row is clustered into the same category, and the classes are novel categories on the evaluation dataset. The left five columns are well-clustered, while the last two are not. The sixth-column images are still reasonable, but the last-columns are the worst cases.

\vspace*{-1mm}
\subsection{Two-step Novel Category Discovery}\label{sec:experiment_twostep}
We present a two-step incremental category discovery experiment on the CUB-200 dataset using ResNet-18. The dataset configurations are more complicated as the datasets are joint sets in incremental steps. The initial step, the first incremental step, and the second incremental step have novel classes in each step, at a rate of $8\colon{1}\colon{1}$, respectively. Each step has its dataset and is indicated as $\mathcal{D}^0$, $\mathcal{D}^1$, and $\mathcal{D}^2$. The samples belonging to novel labeled classes in the initial step are assigned to $\mathcal{D}^0$, $\mathcal{D}^1$, and $\mathcal{D}^2$ at a rate of $8\colon{1}\colon{1}$. Similarly, the samples belonging to novel classes in the first incremental step are assigned to $\mathcal{D}^1$, and $\mathcal{D}^2$ at a rate of $8\colon{2}$. Finally, the rest of the samples are assigned to $\mathcal{D}^2$.

\begin{figure}[t]
    \centering
    \small
    \resizebox{1.\linewidth}{!}{
    \setlength{\tabcolsep}{1.pt}
    \renewcommand{\arraystretch}{0.75}
    \begin{tabular}{cc}
        {\includegraphics[width=1.\linewidth]{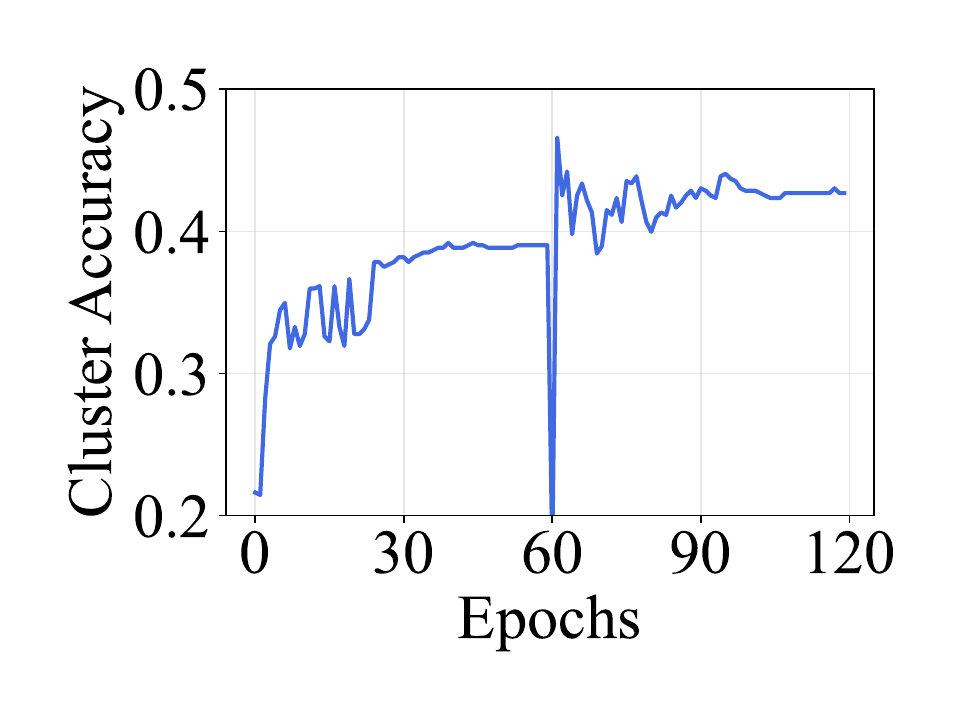}} &{\includegraphics[width=1.\linewidth]{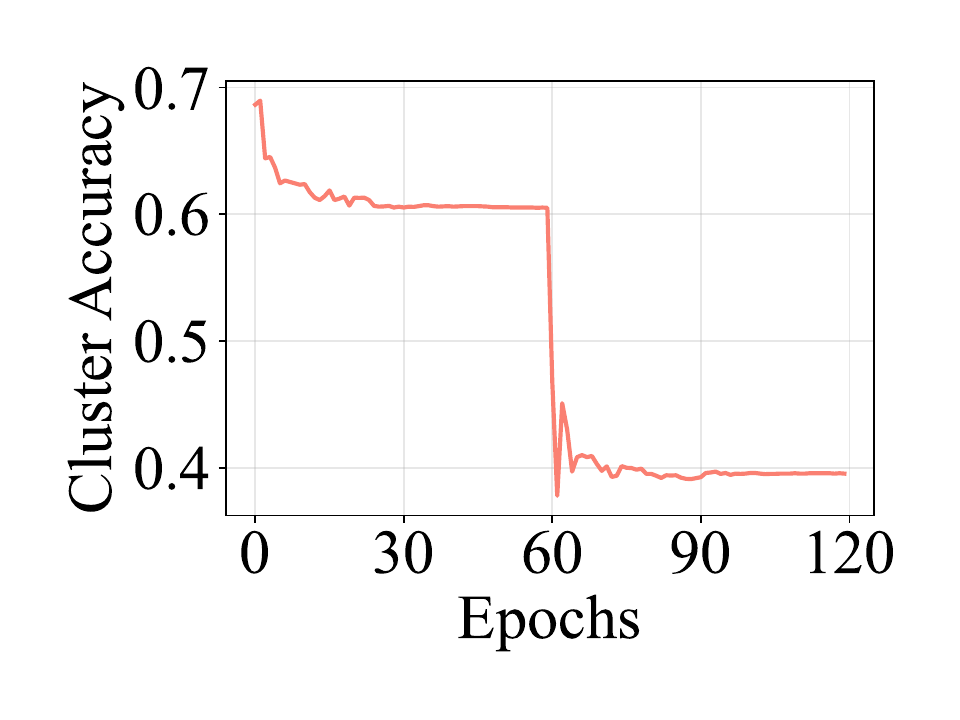}}\\
        \fontsize{0.6cm}{0.6cm}\selectfont{(a) Novel category discovery $M_{d}$} &\fontsize{0.6cm}{0.6cm}\selectfont{(b) Catastrophic forgetting $M_{f}$}\\
    \end{tabular}}
    \caption{Performances of novel category discovery and forgetting on two-step incremental novel categories experiment}
    \label{fig:ablation_twostep}
    \vspace*{-5mm}
\end{figure}

\cref{fig:ablation_twostep} describes the performance of the experiments. Since each incremental step is trained for $60$ epochs, there are deep drops at the $60th$ epoch when new PAs are added. In addition, the cluster accuracy of the old categories, which belong to the $\mathcal{D}^0$, decreases by about $20\%$. The reason is that the number of old categories in the second incremental step is increased compared to the first step. The exemplar cannot focus on only generating the features of $\mathcal{D}^0$. However, the performance of $\mathcal{M}_d$ increased steadily. 

\subsection{Ablation Study}\label{sec:experiment_ablation}
\noindent\textbf{Effectiveness of separations:} We conducted an ablation study to show the effectiveness of our proposed splitter, which consists of a combination of two different methods: the cosine similarity score and a binary splitter using the noise label approach. The ablation experiments were designed such that one used only the cosine similarity score, and another used both methods. As shown in \cref{tab:ablation_networks}, the approach using both methods presents improvements in both old and novel discovery performances. The results reveal the effectiveness of noise labeling for data separation.

\noindent\textbf{Effectiveness of Proxy anchor exemplar:}
To mitigate catastrophic forgetting, various approaches, such as replay~\cite{Bang_2021_CVPR}, prototype~\cite{rebuffi2017icarl, roy2022class}, and pseudo-latents~\cite{joseph2022ncdwf}, have been proposed. Most of these methods exploit the computed average of feature embedding vectors or input data-driven values. However, we propose a novel PA-based exemplar approach and evaluate its efficiency. As shown in \cref{tab:ablation_exempler}, the method without the exemplar recorded the highest novel discovery performances but also showed the highest forgetting. Adopting a general exemplar approach using mean and standard deviation values from the former dataset, $\mathcal{M}_f$ slightly decreased, but $\mathcal{M}_d$ is the lowest. However, our proposed method with the PA-based exemplar recorded the best $\mathcal{M}_f$ and competitive $\mathcal{M}_d$. We analyze that the PAs have representatives of each cluster since PA inherits the relation between data to data and then representative figures of each class. Hence, our PA-based method leads to mitigating forgetting, and we confirm that it is a proper approach.
\begin{table}[t]
\small
\begin{center}
\setlength{\tabcolsep}{1pt}
\renewcommand{\arraystretch}{1.0}
\begin{tabular}{p{0.2\columnwidth}P{0.3\columnwidth}P{0.12\columnwidth}P{0.11\columnwidth}P{0.11\columnwidth}P{0.11\columnwidth}}
    \toprule
    \multirow{2}[2]{*}[-1pt]{Network} &\multirow{2}[2]{*}[-1pt]{Fine Split} &\multicolumn{4}{c}{Metric}\\
    \cmidrule{3-6}
    & &$\mathcal{M}_{all}\uparrow$ &$\mathcal{M}_{o}\uparrow$ &$\mathcal{M}_{f}\downarrow$ &$\mathcal{M}_{d}\uparrow$\\
    \midrule
    \midrule
    \multirow{2}[2]{*}[1pt]{ResNet-18} &without &53.32 &57.63 &16.56 &36.47\\
    &with &54.75 &58.80 &15.47 &40.90\\
    \midrule
    \multirow{2}[2]{*}[1pt]{ResNet-50} &without &66.53 &71.07 &9.12 &48.76\\
    &with &68.09 &71.75 &8.44 &53.79\\
    \midrule
    \multirow{2}[2]{*}[1pt]{ViT-B-16} &without &70.22 &73.02 &10.75 &59.24\\
    &with &72.51 &74.28 &9.49 &65.60\\
    \bottomrule
\end{tabular}
\end{center}
\vspace*{-2mm}
\caption{Ablation study for the proposed fine split on the CUB-200 using three different network architectures, including ResNet-18 and ResNet-50 pre-trained on ImageNet, and ViT-B-16 pre-trained on DINO-ImageNet. The results present the mean over three runs.}
\vspace*{-3mm}
\label{tab:ablation_networks}
\end{table}
\begin{table}[t]
\small
\begin{center}
\setlength{\tabcolsep}{1pt}
\renewcommand{\arraystretch}{1.0}
\begin{tabular}{p{0.2\columnwidth}P{0.3\columnwidth}P{0.12\columnwidth}P{0.11\columnwidth}P{0.11\columnwidth}P{0.11\columnwidth}}
    \toprule
    \multirow{2}[2]{*}[-1pt]{Network} &\multirow{2}[2]{*}[-1pt]{Exemplar} &\multicolumn{4}{c}{Metric}\\
    \cmidrule{3-6}
    & &$\mathcal{M}_{all}\uparrow$ &$\mathcal{M}_{o}\uparrow$ &$\mathcal{M}_{f}\downarrow$ &$\mathcal{M}_{d}\uparrow$\\
    \midrule
    \midrule
    \multirow{3}[2]{*}[1pt]{ResNet-18} &without &38.24 &37.31 &36.88 &41.89\\
    &Data mean and std. &40.23 &40.56 &33.63 &37.22\\
    &Proxy anchors &54.75 &58.80 &15.47 &40.90\\
    \bottomrule
\end{tabular}
\end{center}
\vspace*{-1mm}
\caption{Effectiveness of the proposed PA-based exemplar on the CUB-200 dataset using ResNet-18 pre-trained on ImageNet}
\label{tab:ablation_exempler}
\vspace*{-5mm}
\end{table}
\begin{table}[t]
\small
\begin{center}
\setlength{\tabcolsep}{1pt}
\renewcommand{\arraystretch}{1.0}
\begin{tabular}{P{0.2\columnwidth}P{0.01\columnwidth}P{0.18\columnwidth}P{0.18\columnwidth}P{0.18\columnwidth}P{0.18\columnwidth}}
    \toprule
    {Ratio} & &\multicolumn{4}{c}{Metric}\\
    \cmidrule{1-1}\cmidrule{3-6}
    $\mathcal{Y}_{old}\colon\mathcal{Y}_{new}$ & &{$\mathcal{M}_{all}\uparrow$} &{$\mathcal{M}_o\uparrow$} &{$\mathcal{M}_f\downarrow$} &{$\mathcal{M}_d\uparrow$}\\
    \midrule
    \midrule
    {${9}\colon{1}$} & &61.34 &63.22 &11.83 &44.73\\
    {${8}\colon{2}$} & &54.75 &58.80 &15.47 &40.90\\
    {${7}\colon{3}$} & &52.66 &58.42 &14.67 &39.50\\
    {${6}\colon{4}$} & &48.78 &57.90 &15.16 &35.53\\
    {${5}\colon{5}$} & &45.28 &62.41 &11.30 &28.55\\
    \bottomrule
\end{tabular}
\end{center}
\vspace*{-1mm}
\caption{Qualitative evaluation by changing the ratio of classes and samples on the CUB-200 using ResNet18 pre-trained on ImageNet}
\label{tab:ablation_class_ratio}
\vspace*{-5mm}
\end{table}

\noindent\textbf{Robustness of class and sample blending ratio variants:}
In general, the capability to recognize novel categories largely depends on a powerful and well-trained initial model with the target datasets. The more classes and samples are included in the initial training dataset $\mathcal{D}^0$, the better the model learns representative features to fit the sets. To evaluate the robustness of the proposed model, we conducted experiments with variants of the number of samples and classes in $\mathcal{D}^0$. As described in \cref{tab:ablation_class_ratio}, decreasing the number of classes and data in the labeled set decreased the discovery of novel classes and clustering accuracies as the number of unlabeled novel data increased. On the other hand, catastrophic forgetting could increase since the number of novel data increases. However, forgetting was maintained within a reasonable boundary, indicating the effectiveness of our PA-based exemplar. The results suggest that our method has the robustness of the variants.

\vspace*{-1mm}
\section{Related work}\label{sec:related}
\subsection{Novel Category Discovery}\label{sec:related_ncd}
NCD techniques have been proposed to classify data with various constraints on unlabeled data.
One category of the methods presented pre-training the model on the labeled set and fine-tuning it on the unlabeled set using unsupervised clustering losses~\cite{Hsu_2018_ICLR, Hsu_2019_ICLR, Han_2019_ICCV, Liu_2022_TNNLS, 9690577}.
Another category assumed the availability of both the labeled and unlabeled data, and trained networks jointly with a labeled novel class loss within the semi-supervised scheme~\cite{han2019automatically,  Zhong_2021_CVPR_NCL, Zhong_2021_CVPR_openMix, Jia_2021_ICCV, Fini_2021_ICCV, zhao21novel}. 
Han~\etal~\cite{han21autonovel} proposed transferring knowledge from labeled to unlabeled data using ranking statistics in the joint learning stage. 
Recently, GCD~\cite{vaze2022gcd} and XCon~\cite{Fei_2022_BMVC} tackled the more realistic scenario of joint datasets and distinguished known and unknown classes using prior knowledge.
However, these approaches did not consider the continual learning scheme.
To address the limitation, FRoST~\cite{roy2022class} and NCDwF~\cite{joseph2022ncdwf} froze feature extractors and added the second head for each novel class, as much as the given number of novel categories.
However, the methods employed disjoint sets.
GM~\cite{zhang2022grow} proposed to consider novel category discovery on the joint datasets,
but still require prior knowledge, such as the proportion of novel samples.

\subsection{Image Retrieval}\label{sec:related_ir}
Most of the image retrieval methods have utilized metric learning and can be categorized into two approaches. Pair-based methods exploited contrastive loss~\cite{NIPS1993_288cc0ff, 1467314, 1640964} and triplet loss~\cite{schroff2015facenet, wang2014learning}, that pull together data pairs in the same class and push apart those in different classes. Multiple data-based~\cite{NIPS2016_6b180037, oh2016deep} methods proposed considering the relations between multiple data. Entire data-based approaches~\cite{wang2019ranked, wang2019multi} presented considering all data in a batch, leveraging fine-grained semantic relations between them while requiring high computation costs and slow convergence. In contrast, proxy-based methods~\cite{movshovitz2017no, qian2019softtriple, aziere2019ensemble} employed fewer proxies than the training set, reducing training complexity. While these methods improved training convergence, they did not consider data-to-data relations, as each data was associated with its proxy. PA~\cite{Kim_2020_CVPR} inherited the strength of pair- and proxy-based methods, achieving fast and reliable convergence, robustness opposing noisy data, and leveraging rich data-to-data relations.

\subsection{Noise Label}\label{sec:related_nl}
Recently proposed methods for learning with noisy labels have highlighted the importance of discriminating between clean and noise-labeled data to improve performance. DivideMix~\cite{li2020dividemix} used GMM to distinguish between clean and noisy labeled data and treated the latter as unlabeled for semi-supervised learning. AugDesc~\cite{nishi2021augmentation} employed data augmentation to enhance the differentiation between clean and noisy labeled data, while INCV~\cite{chen2019understanding} introduced cross-validation to separate clean data from noisy training data. SplitNet~\cite{kim2022splitnet} leveraged a compact network to perceive the difference between clean and noisy labels, improving model performance by more accurately differentiating noise. 

\vspace*{-3mm}
\section{Conclusion}\label{sec:conclusion}
In this paper, we presented a novel continual learning scenario, considered NCD on the unlabeled joint datasets without any prior knowledge of the dataset. Our framework utilized PAs to split known and novel categories, resulting in well-clustered and well-pseudo-labeled categories that mitigate catastrophic forgetting. We further refined the splitting of the dataset by adopting a noise labeling scheme. Our proposed approach outperformed existing state-of-the-art methods regarding novel category discovery and forgetting. While DeepDPM~\cite{Ronen_2022_CVPR} has recently shown outstanding performance on non-parametric clustering tasks, we believe that our proposed method can achieve even better performance by adopting better clustering manners. In future work, we plan to evaluate our method by adopting a better clustering manner.

\vspace*{-3mm}
\section*{Acknowledgements}
This work was funded by Samsung Electro-Mechanics and was partially supported by Carl-Zeiss Stiftung under the Sustainable Embedded AI project (P2021-02-009).

{\small
\bibliographystyle{ieee_fullname}
\bibliography{egbib}
}

\renewcommand{\thetable}{\Alph{table}}
\renewcommand{\thefigure}{\Alph{figure}}
\renewcommand{\thealgorithm}{\Alph{algorithm}}

\appendix
\clearpage
\setcounter{table}{0}
\setcounter{figure}{0}

\section{Pseudo Code}\label{sec:code}
\begin{algorithm}[!t]
    \centering
	\caption{Pseudo-code of the one-step incremental novel category discovering} 
	\begin{algorithmic}[1]
        \State\textbf{Initial step:}
        \State\quad{Given labeled dataset $\mathcal{D}^0=\{(x, y)\}$}
        \State\quad{Train network $f^0(\cdot)$ and proxy anchors $g^0(\cdot)$}
        \State\quad{Calculate $\sigma$ of $\mathcal{D}^0$ for exemplar $\mathcal{E}^0$}
        \State\textbf{Discovering novel category step:}
        \State\quad{Given unlabeled joint dataset $\mathcal{D}^1=\{x\}$}
        \State\quad{Extract embedding vectors $z_i$ on $f^0(x_i)$}
        \State\quad{\parbox[t]{\dimexpr\linewidth-\algorithmicindent}{Get initial separated datasets on measuring cosine similarity score $s(z_i, p)$ using the initial split}}
        \State\quad{\parbox[t]{\dimexpr\linewidth-\algorithmicindent}{Get fine separated datasets, $\mathcal{D}^1_{old}=\{x^1_{old}\}$ and $\mathcal{D}^1_{new}=\{x^1_{new}\}$ on Noisy labeling and Gaussian mixture model using the initial separated dataset}}
        \State\quad{\parbox[t]{\dimexpr\linewidth-\algorithmicindent}{Get pseudo-labels $\mathcal{D}^1_{old}$ using previous network $f^0$, $\mathcal{\hat{D}}^1_{old}=\{(x^1_{old}, \hat{y}^1_{old}=\argmax(f^0(x^1_{old}))\}$}}
        \State\quad{\parbox[t]{\dimexpr\linewidth-\algorithmicindent}{Get pseudo-labels $\mathcal{D}^1_{new}$ using a non-parametric clustering approach $c(\cdot)$, affinity propagation, $\mathcal{\hat{D}}^1_{new}=\{(x^1_{new}, \hat{y}^1_{new}=\argmax(c(x^1_{new}))\}$}}
        \State\textbf{Category incremental step:}
        \State\quad{\parbox[t]{\dimexpr\linewidth-\algorithmicindent}{Add new proxy anchors as the estimated number of novel categories based on $c(\cdot)$ results}}
        \State\quad{\parbox[t]{\dimexpr\linewidth-\algorithmicindent}{Assign initial means of newly added proxy anchors based on $c(\cdot)$ results}}
        \State\quad{Generate old embedding vectors $\tilde{z}^0$ using $\mathcal{E}^0$}
        \State\quad{\parbox[t]{\dimexpr\linewidth-\algorithmicindent}{Train network $f^1(\cdot)$ and modified Proxy anchors $g^1(\cdot)$ using pseudo-labeled dataset $\mathcal{\hat{D}}^1$ and $\tilde{z}^0$}}
        \State\quad{\parbox[t]{\dimexpr\linewidth-\algorithmicindent}{Distill knowledge between networks, $f^0(\cdot)$ and $f^1(\cdot)$}}
        \State\quad{Calculation $\sigma$ of $\mathcal{\hat{D}}^1$ for exemplar $\mathcal{E}^1$}
	\end{algorithmic}
\label{algo:algo}
\end{algorithm}

Pseudo-code of our proposed method is represented in Algorithm~\ref{algo:algo} in detail. The code is written for the one-time step procedure and is comprised of three steps: the initial step, the discovering novel category step, and the category incremental step. The initial step is fine-tuning the network on the dataset and training proxy anchors using the labeled dataset. Then, the following given dataset for incremental category learning is the unlabeled joint set, which includes the old and novel classes. In the discovering novel category step, we separate the dataset into old and novel categories using the initial split and the fine split, and pseudo-label the separated datasets. In the last step, exploiting the pseudo-labeled dataset, we add new proxy anchors and fine-tune the network and proxy anchors. Alleviating the catastrophic forgetting, we utilize proxy anchor-based exemplar and feature distillation.

For extending to continuously incremental novel categories, the initial step is trained only once, then the discovering novel category step and the category incremental step are trained iteratively and sequentially.
Specifically, let us assume that the novel category discovery continually increases until the $n\nth$ step. In the discovering novel category step, the given dataset notation is changed from $\mathcal{D}^1$ to $\mathcal{D}^n$, and the previous network $f^0$ is also replaced to $f^{n-1}$. Also, the pseudo-labeled dataset $\hat{\mathcal{D}^1}$ is modified to $\hat{\mathcal{D}^n}$. In the category incremental step, the generated vector $\tilde{z}^0$, the exploited exemplar $\mathcal{E}^0$ for the vector, and the pseudo-labeled dataset $\hat{\mathcal{D}^1}$ are notated $\tilde{z}^{n-1}$, $\mathcal{E}^{n-1}$, and $\hat{\mathcal{D}^n}$, respectively. $f^1(\cdot)$ and $g^1(\cdot)$ are replaced $f^n(\cdot)$ and $g^n(\cdot)$. Lastly, the new exemplar $\mathcal{E}^1$ is substituted with $\mathcal{E}^n$.
\begin{figure*}[t]
    \centering
    \resizebox{1.\linewidth}{!}{
    \setlength{\tabcolsep}{1pt}
    \begin{tabular}{cccc}
        \includegraphics[width=1\linewidth]{figures/Cos_similarity_1.pdf} &\includegraphics[width=1\linewidth]{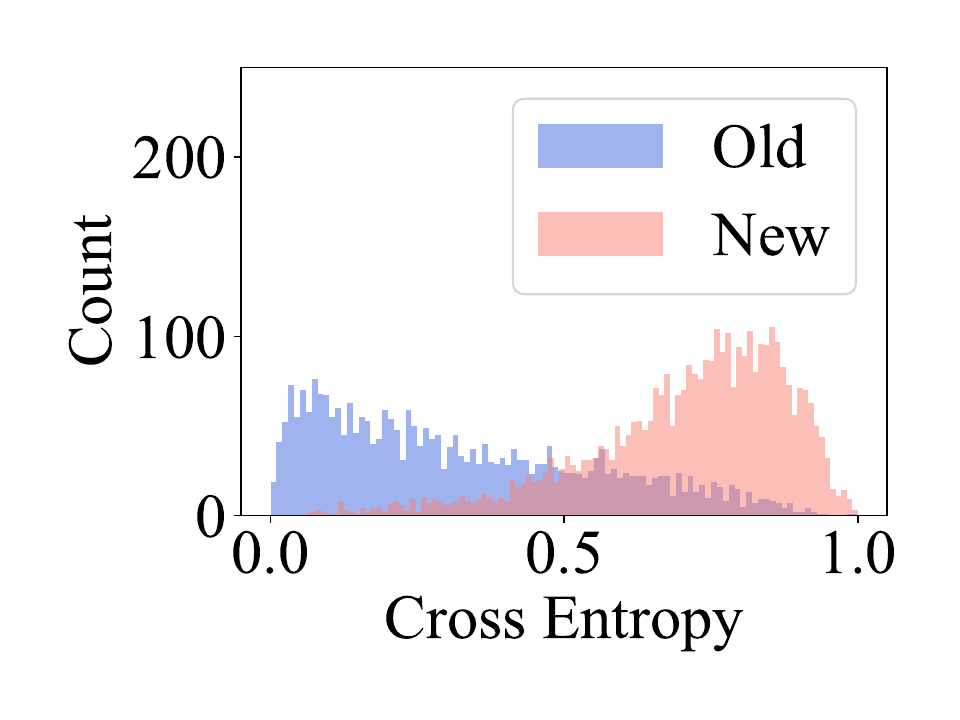}        &\includegraphics[width=1\linewidth]{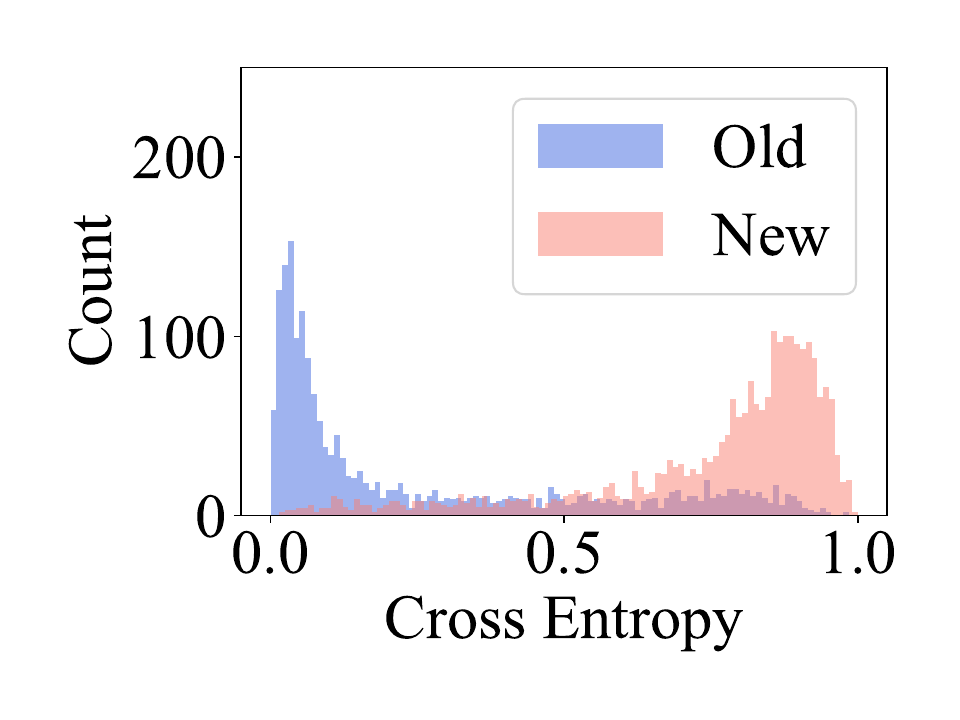}        &\includegraphics[width=1\linewidth]{figures/cross_entropy_1_1.pdf}\\
        \fontsize{1.3cm}{1.3cm}\selectfont{(a) Initial Split} &\fontsize{1.3cm}{1.3cm}\selectfont{(b) Fine Split: the first epoch} &\fontsize{1.3cm}{1.3cm}\selectfont{(c) Fine Split: the second epoch} &\fontsize{1.3cm}{1.3cm}\selectfont{(d) Fine Split: the last epoch}
    \end{tabular}}
    \vspace*{1mm}
    \caption{The initial split and the fine split results using CUB-200 dataset on ResNet-18. The given dataset is joint unlabeled, and we separate the dataset into old and novel categories without any prior knowledge}
    \label{fig:sup_fine_split}
\end{figure*}
\begin{table*}[t!]
\small
\begin{center}
\setlength{\tabcolsep}{1pt}
\renewcommand{\arraystretch}{1.0}
\begin{tabular}{p{0.2\linewidth}p{0.07\linewidth}p{0.12\linewidth}P{0.08\linewidth}P{0.08\linewidth}P{0.08\linewidth}P{0.08\linewidth}P{0.08\linewidth}P{0.08\linewidth}P{0.08\linewidth}}
    \toprule
    Method &Step &Novel Classes &$\mathcal{M}_{all}\uparrow$ &$\mathcal{M}_{o}^0\uparrow$ &$\mathcal{M}_{f}\downarrow$ &$\mathcal{M}_{n}^1\uparrow$ &$\mathcal{M}_{n}^2\uparrow$ &$\mathcal{M}_{n}^3\uparrow$ &$\mathcal{M}_{d}\uparrow$\\
    \midrule
    \midrule
    \multirow{6}[0]{*}[-1pt]{GM-CI~\cite{zhang2022grow}} &$0\nth$ &$0\sim{139}$ &59.51 &59.51 &- &- &- &- &-\\
    &$1\st$ &$140\sim{159}$ &13.55 &13.48 &46.03 &13.99 &- &- &13.99\\
    &$2\nd$ &$160\sim{179}$ &37.32 &41.36 &18.15 &25.43 &21.62 &- &23.53\\
    &$3\rd$ &$180\sim{199}$ &35.74 &40.44 &19.07 &27.13 &19.93 &28.06 &25.04\\
    \cmidrule{2-10}
    &$\ast$ & &- &- &19.48 &- &- &- &24.97\\
    \cmidrule{2-10}
    &$0\nth$ &$0\sim{159}$ &60.34 &60.34 &- &- &- &- &-\\
    &$1\st$ &$160\sim{199}$ &7.13 &7.28 &53.06 &6.53 &- &- &6.53\\
    \midrule
    \multirow{6}[0]{*}[-1pt]{GM-MI~\cite{zhang2022grow}} &$0\nth$ &$0\sim{139}$ &26.54 &26.54 &- &- &- &- &-\\
    &$1\st$ &$140\sim{159}$ &18.70 &20.48 &6.06 &7.51 &- &- &7.51\\
    &$2\nd$ &$160\sim{179}$ &19.90 &23.24 &3.30 &6.83 &10.14 &- &8.49\\
    &$3\rd$ &$180\sim{199}$ &17.22 &20.80 &5.74 &6.66 &8.28 &12.24 &6.89\\
    \cmidrule{2-10}
    &$0\nth$ &$0\sim{159}$ &46.39 &46.39 &- &- &- &- &-\\
    &$1\st$ &$160\sim{199}$ &\textbf{6.43} &\textbf{6.57} &\textbf{39.82} &\textbf{5.92} &- &- &\textbf{5.92}\\
    \midrule
    \multirow{2}[0]{*}[-1pt]{\textbf{Ours-CGCD}} &$0\nth$ &$0\sim{159}$ &74.27 &74.27 &- &- &- &- &-\\
    &$1\st$ &$160\sim{200}$ &\textbf{54.75} &\textbf{58.80} &\textbf{15.47} &\textbf{40.90} &- &- &\textbf{40.90}\\
    \bottomrule
\end{tabular}
\end{center}
\caption{Comparison of GM~\cite{zhang2022grow} method evaluation results on the two different scenarios, which are Class incremental scenario (CI) and Mixed Incremental Scenario (MI) are proposed originally in GM paper. The experiment results exploited the CUB-200 dataset on ResNet-18. For a fair comparison, we followed the hyperparameters on their report. Nevertheless, GM is underperformed significantly. $\ast$ denotes results reported in the original GM paper and the bold results indicated the reported in the main paper.}
\vspace*{-3mm}
\label{tab:sup_problem_sota}
\end{table*}
Our code is made available at \url{https://github.com/Hy2MK/CGCD}.
\section{Fine Split}\label{sec:fine}
To acquire a more clearly separated dataset without noisy data, we design a simple network for the fine split. The network comprises a series of the Fully Connected layer (FC) - Batch Normalization (BN) - sigmoid - FC - BN - sigmoid - FC. The data is selected on both ends of the spectrum for training the network since we assume the data in the region is clean 
(\ie lower than $5\%$ and over $95\%$ based on the result of GMM).
The simple perceptron model aims to predict whether the noisy data belongs to the old or novel categories and is trained on three epochs. 
\cref{fig:sup_fine_split} (a) is depicted the initial split result using cosine similarity score measurement between the embedding vectors and proxy anchors. There is an overlapped region between the old and novel categories, which represents the initial split using previous knowledge is unclear to divide novel and old categories. Therefore, we adopt the noisy labeling scheme to fine separation into old and novel classes, and confirmed the results of every epoch of the fine split from \cref{fig:sup_fine_split} (b) to \cref{fig:sup_fine_split} (d).

\section{GM Results Analysis}\label{sec:gm}
In the paper, we conducted the comparison experiment of our proposed method with state-of-the-art approaches. Among the compared methods, GM~\cite{zhang2022grow} is recorded as significantly underperforming. Therefore, we should confirm to clear that our experiments are conducted in a fair comparison following the hyperparameters reported in the original paper without any modifications.

GM proposed four different scenarios, which are Class Incremental scenario (CI), Data Incremental Scenario (DI), Mixed Incremental scenario (MI), and Semi-supervised Mixed Incremental Scenario (SMI). Among the scenarios, CI and MI are the most similar to ours, Continuous Generalized Category Discovery (CGCD). We evaluated these two scenarios using CUB-200 dataset on ResNet-18. As presented in~\cref{tab:sup_problem_sota}, GM conducted experiments using fine-grained datasets only in the CI scenario, and in the other scenario, the CIFAR-100 dataset was utilized to evaluate the performance of the method. In this sense, we confirmed the reproducibility in the CI scenario using CUB-200 dataset since the $\mathcal{M}_d$ and $\mathcal{M}_f$ results were equivalent.
The MI scenario is the closest to our proposed scenario CGCD. However, GM presented significant underperformance in the MI scenarios, and $\mathcal{M}_d$ performances were recorded at under $10\%$, particularly in both three-time incremental and one-time incremental scenarios.

On the other hand, we confirmed outstanding performance compared to GM in this experiment, and our proposed method also recorded improved performance in the two-step incremental scenario reported in the paper.

Following the dataset policy of GM, their given distribution is $7:1:1:1$, and the results using the original parameters reported in the paper are shown in Table~\ref{tab:experiment_gm}. As the step increases, $\mathcal{M}_{f}$ and $\mathcal{M}_{d}$ recorded mean values of $24.01$ and $14.92$, respectively.
But, in our ablations, we evaluated the policy, $7:3$. This means that we classify into joint unlabeled novel data more than three times as GM at once. Through experiments, we implicitly showed superior performance.
\begin{table}[t]
\footnotesize
\begin{center}
\setlength{\tabcolsep}{1pt}
\renewcommand{\arraystretch}{1.0}
\begin{tabular}{p{0.11\columnwidth}P{0.17\columnwidth}P{0.1\columnwidth}P{0.1\columnwidth}P{0.01\columnwidth}P{0.1\columnwidth}P{0.1\columnwidth}P{0.01\columnwidth}P{0.1\columnwidth}P{0.1\columnwidth}}
    \toprule
    \multirow{2}[2]{*}[-1pt]{~Method} &\multirow{2}[2]{*}[-1pt]{~Dataset} &\multicolumn{2}{c}{$1^{st}$} & &\multicolumn{2}{c}{$2^{nd}$} & &\multicolumn{2}{c}{$3^{rd}$}\\
    \cmidrule{3-4}\cmidrule{6-7}\cmidrule{9-10}
    & & {$\mathcal{M}_{f}\downarrow$} &{$\mathcal{M}_{d}\uparrow$} & &{$\mathcal{M}_{f}\downarrow$} &{$\mathcal{M}_{d}\uparrow$} & &{$\mathcal{M}_{f}\downarrow$} &{$\mathcal{M}_{d}\uparrow$}\\
    \midrule
    \midrule
    ~GM &CUB-200 &30.99 &12.21 & &19.86 &15.99 & &21.18 &16.56\\
    \bottomrule
\end{tabular}
\end{center}
\caption{GM step-wise results following GM's original process.}
\label{tab:experiment_gm}
\end{table}

\section{Adopting other deep metric learning}\label{sec:metric_learn}
\begin{table}[h]
\footnotesize
\begin{center}
\setlength{\tabcolsep}{1pt}
\renewcommand{\arraystretch}{1.0}
\begin{tabular}{p{0.25\columnwidth}P{0.18\columnwidth}P{0.18\columnwidth}P{0.18\columnwidth}P{0.18\columnwidth}}
    \toprule
    \multirow{2}[2]{*}[-1pt]{Method} &\multicolumn{4}{c}{CUB-200}\\
    \cmidrule{2-5}
    & {$\mathcal{M}_{all}\downarrow$} &{$\mathcal{M}_{o}\uparrow$} &{$\mathcal{M}_{f}\downarrow$} &{$\mathcal{M}_{d}\uparrow$}\\
    \midrule
    \midrule
    ArcFace~\cite{deng2019arcface} &53.13$\pm$2.76 &60.21$\pm$3.80 &13.52$\pm$3.81 &28.02$\pm$1.39\\
    ProxyNCA++~\cite{teh2020proxynca++} &53.72$\pm$0.85 &\textbf{65.62$\pm$0.52} &\textbf{8.85$\pm$0.52} &25.00$\pm$1.77\\
    \textbf{Ours} &\textbf{54.75$\pm$0.64} &{58.80$\pm$0.99} &{15.47$\pm$0.99} &\textbf{40.90$\pm$1.07}\\
    \bottomrule
\end{tabular}
\end{center}
\caption{Ablation study for adopting various deep metric learning.}
\label{tab:experiment_metric}
\end{table}

Table~\ref{tab:experiment_metric} shows the results of replacing PAs with others.~\cite{deng2019arcface, teh2020proxynca++}.
We assumed PAs is trained in more valuable features for utilizing to split novel and old category since having both proxy- and anchor-based merits. As well-separated novel data samples increased, the result showed $\mathcal{M}_{f}$ decreased, but $\mathcal{M}_{d}$ improved. And we confirmed PAs is to fit our framework.

\section{Ours Qualitative Results}\label{sec:qual}
\begin{figure*}[t]
    \centering
    \small
    \resizebox{1\linewidth}{!}{
    \setlength{\tabcolsep}{1.pt}
    \begin{tabular}{ccccccc}
        {\includegraphics[width=1.\linewidth]{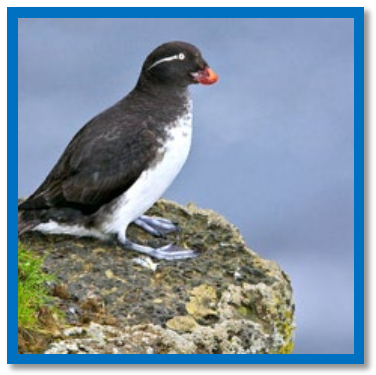}} &{\includegraphics[width=1.\linewidth]{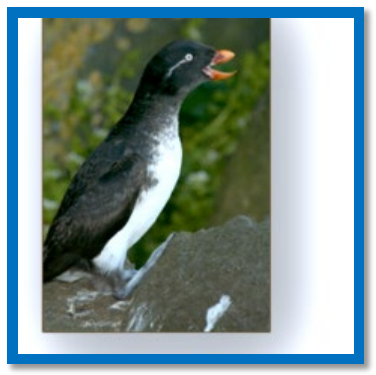}} &{\includegraphics[width=1.\linewidth]{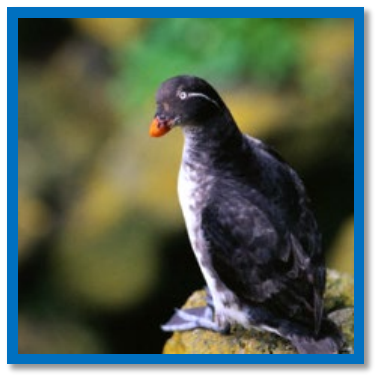}} &{\includegraphics[width=1.\linewidth]{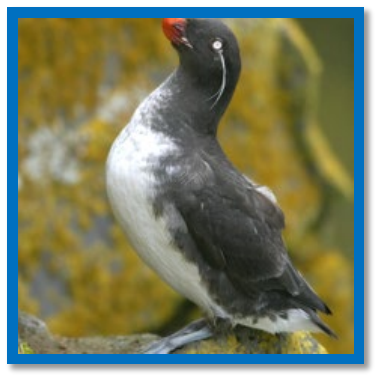}} &{\includegraphics[width=1.\linewidth]{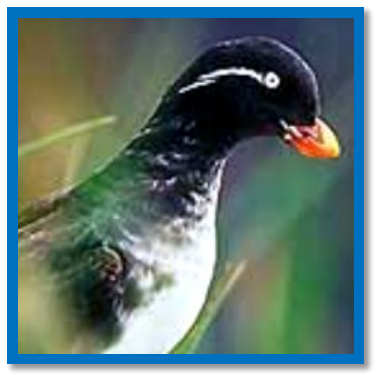}} &{\includegraphics[width=1.\linewidth]{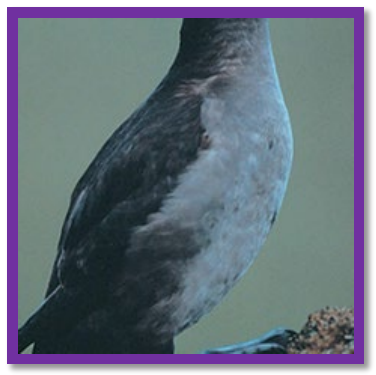}} &{\includegraphics[width=1.\linewidth]{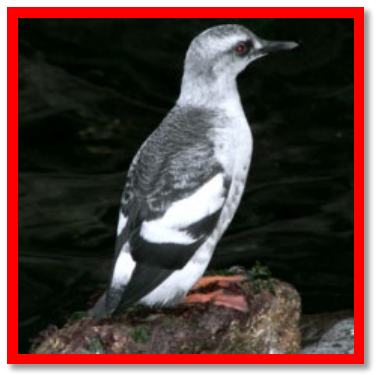}}\\
        \midrule[10pt]
        {\includegraphics[width=1.\linewidth]{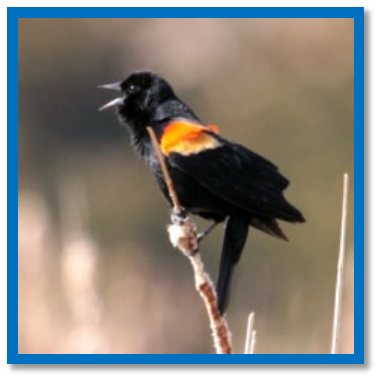}} &{\includegraphics[width=1.\linewidth]{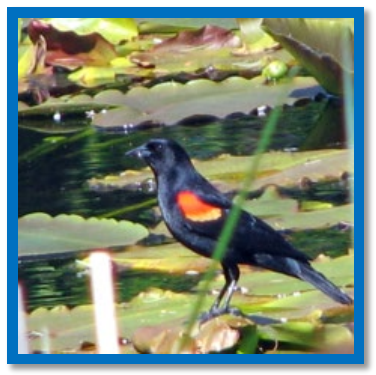}} &{\includegraphics[width=1.\linewidth]{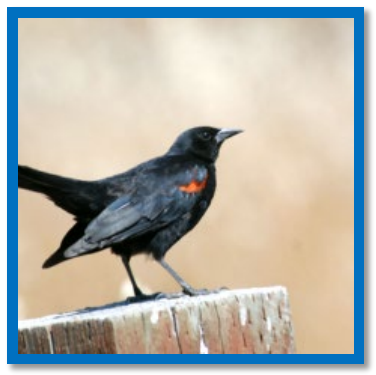}} &{\includegraphics[width=1.\linewidth]{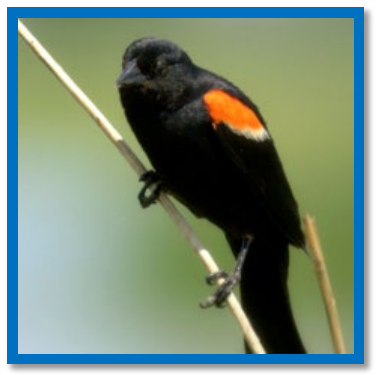}} &{\includegraphics[width=1.\linewidth]{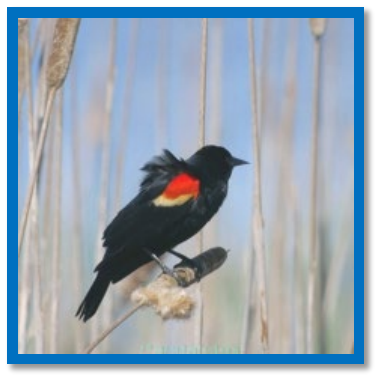}} &         {\includegraphics[width=1.\linewidth]{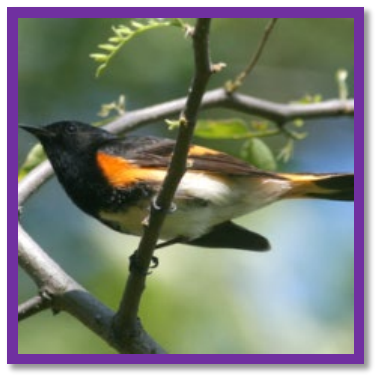}} & {\includegraphics[width=1.\linewidth]{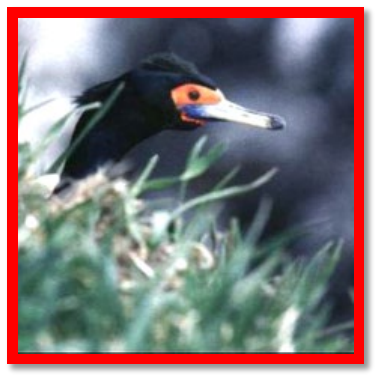}}\\
        \midrule[10pt]
        {\includegraphics[width=1.\linewidth]{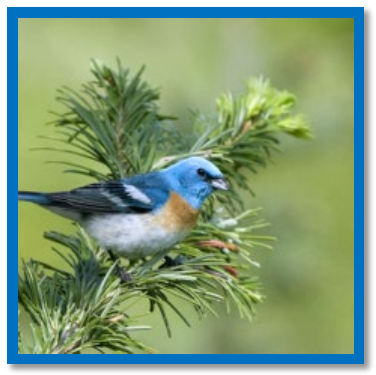}} &{\includegraphics[width=1.\linewidth]{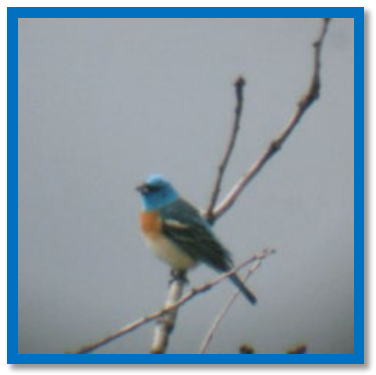}} &{\includegraphics[width=1.\linewidth]{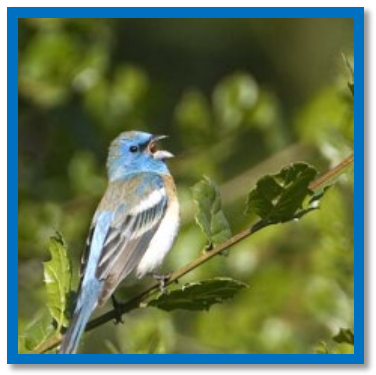}} &{\includegraphics[width=1.\linewidth]{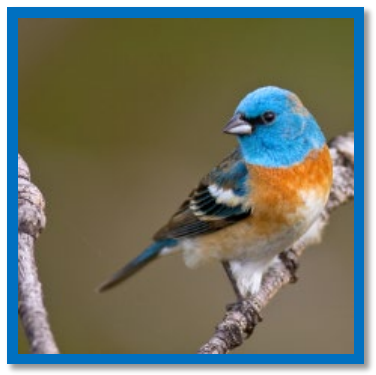}} &{\includegraphics[width=1.\linewidth]{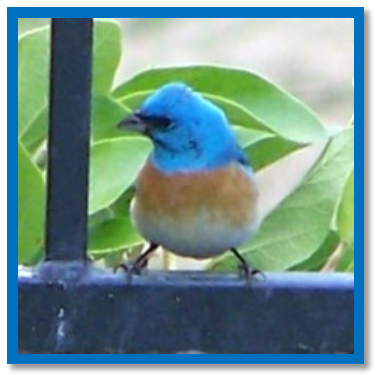}} &         {\includegraphics[width=1.\linewidth]{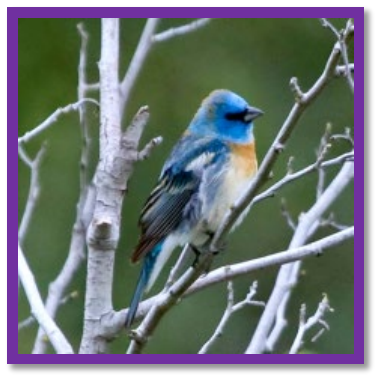}} & {\includegraphics[width=1.\linewidth]{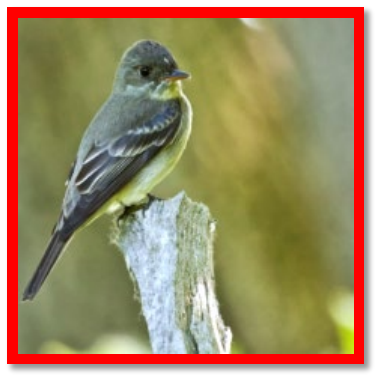}}\\
        \midrule[10pt]
        {\includegraphics[width=1.\linewidth]{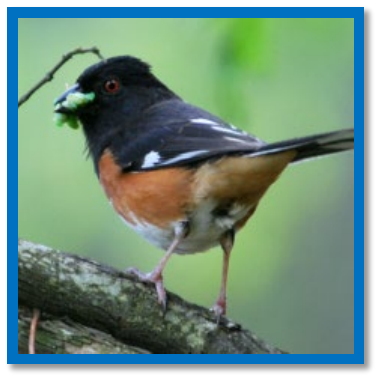}} &{\includegraphics[width=1.\linewidth]{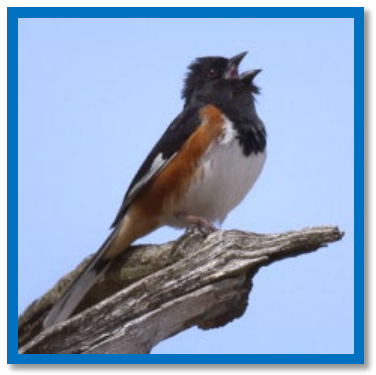}} &{\includegraphics[width=1.\linewidth]{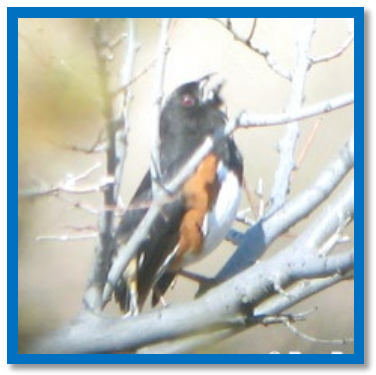}} &{\includegraphics[width=1.\linewidth]{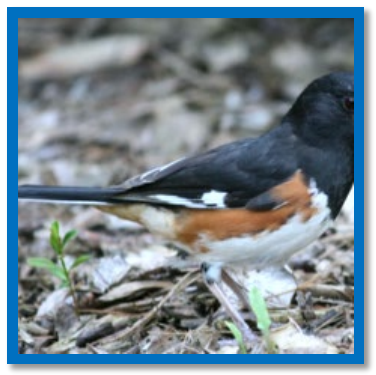}} &{\includegraphics[width=1.\linewidth]{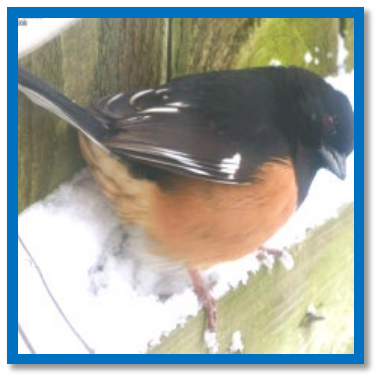}} &         {\includegraphics[width=1.\linewidth]{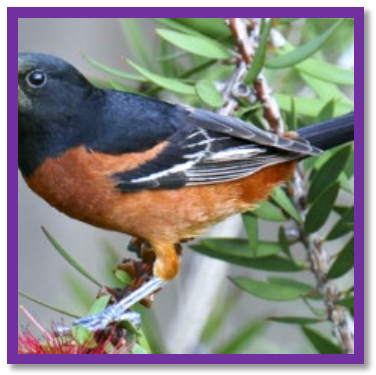}} & {\includegraphics[width=1.\linewidth]{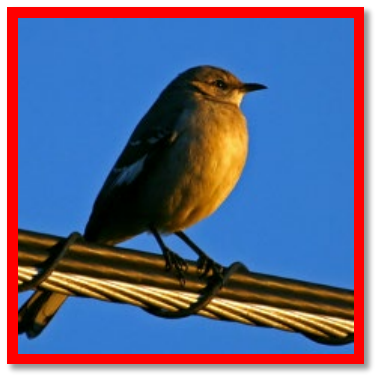}}\\
        \midrule[10pt]
        {\includegraphics[width=1.\linewidth]{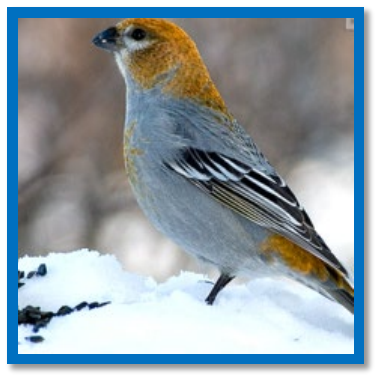}} &{\includegraphics[width=1.\linewidth]{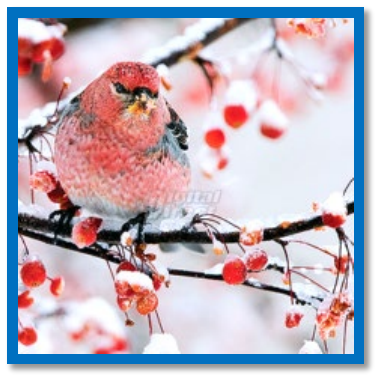}} &{\includegraphics[width=1.\linewidth]{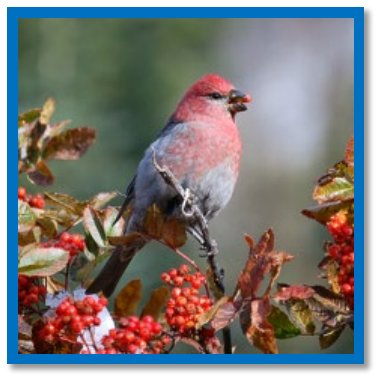}} &{\includegraphics[width=1.\linewidth]{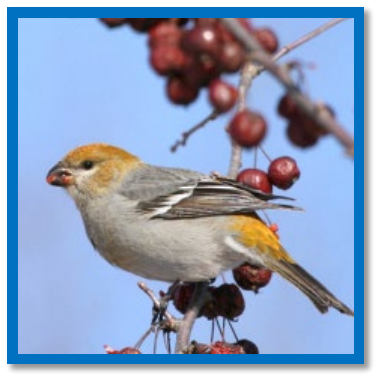}} &{\includegraphics[width=1.\linewidth]{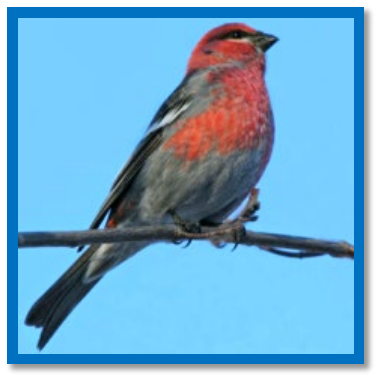}} &         {\includegraphics[width=1.\linewidth]{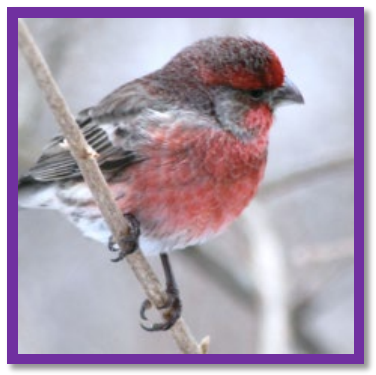}} & {\includegraphics[width=1.\linewidth]{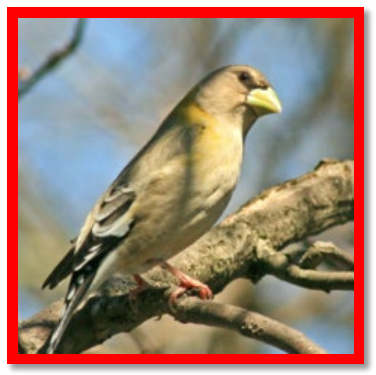}}\\
        \midrule[10pt]
        {\includegraphics[width=1.\linewidth]{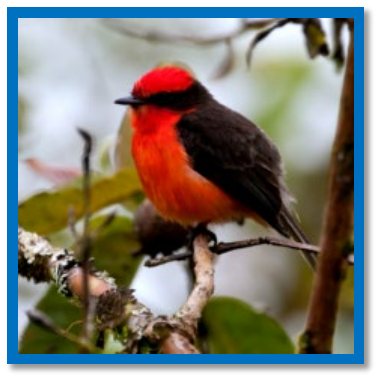}} &{\includegraphics[width=1.\linewidth]{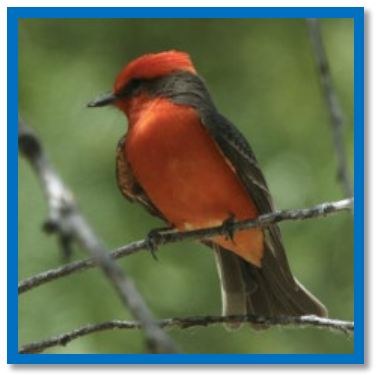}} &{\includegraphics[width=1.\linewidth]{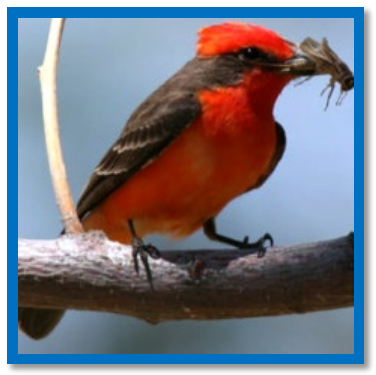}} &{\includegraphics[width=1.\linewidth]{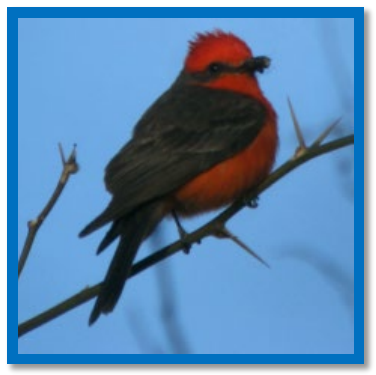}} &{\includegraphics[width=1.\linewidth]{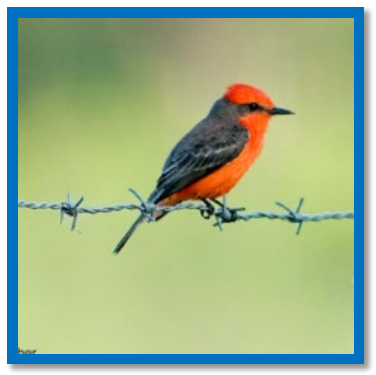}} &         {\includegraphics[width=1.\linewidth]{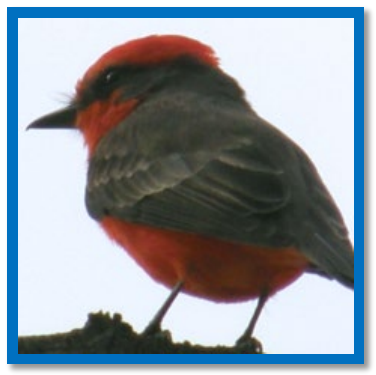}} & {\includegraphics[width=1.\linewidth]{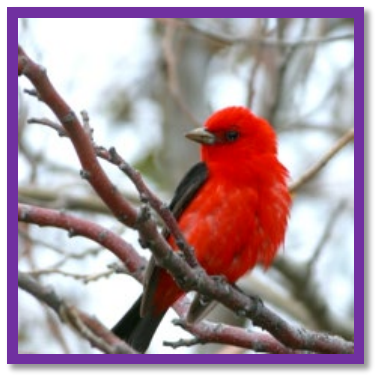}}\\
        \midrule[10pt]
        {\includegraphics[width=1.\linewidth]{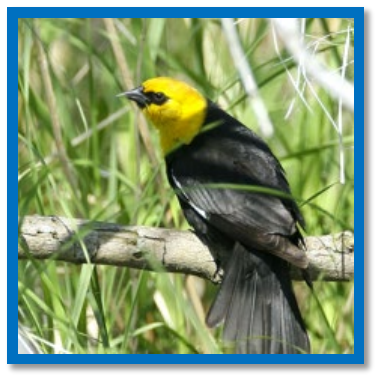}} &{\includegraphics[width=1.\linewidth]{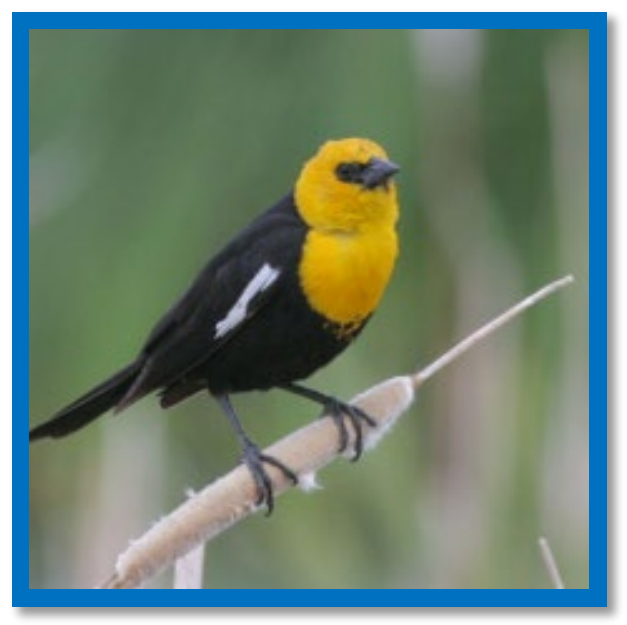}} &{\includegraphics[width=1.\linewidth]{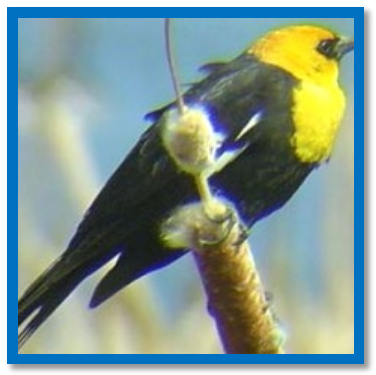}} &{\includegraphics[width=1.\linewidth]{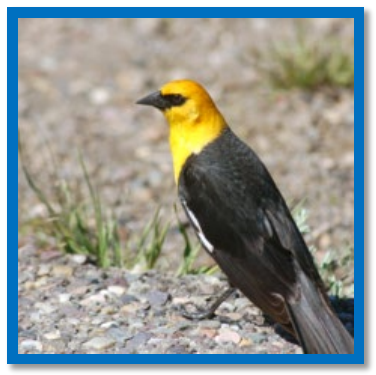}} &{\includegraphics[width=1.\linewidth]{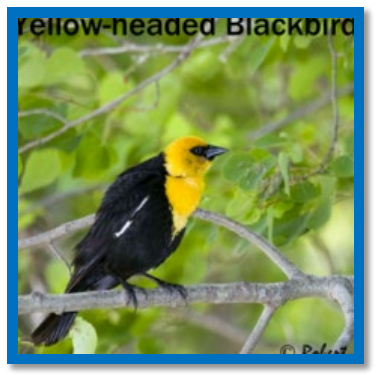}} &         {\includegraphics[width=1.\linewidth]{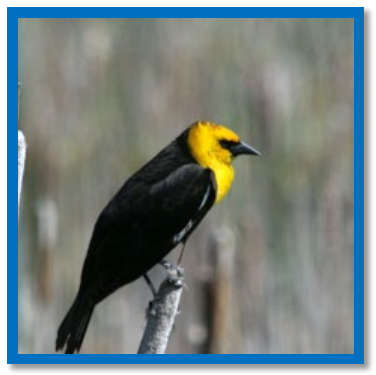}} & {\includegraphics[width=1.\linewidth]{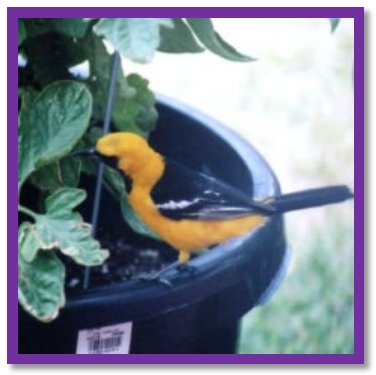}}\\
    \end{tabular}}
    \caption{Qualitative experiment results of the proposed method. The evaluation is conducted using the CUB-200 dataset on ResNet-18, and the images belong to the old categories and are also clustered in the old categories. The first five columns with blue boxes denote well-clustered examples. The last two columns represent failed prediction results, including example images with purple boxes denoting hard negatives and those with red boxes indicating incorrect categorization.}
    \label{fig:experiment_quality_old}
\end{figure*}

\begin{figure*}[t]
    \centering
    \small
    \resizebox{1\linewidth}{!}{
    \setlength{\tabcolsep}{1.pt}
    \begin{tabular}{ccccccc}
        {\includegraphics[width=1.\linewidth]{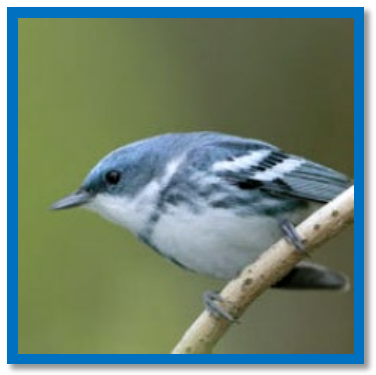}} &{\includegraphics[width=1.\linewidth]{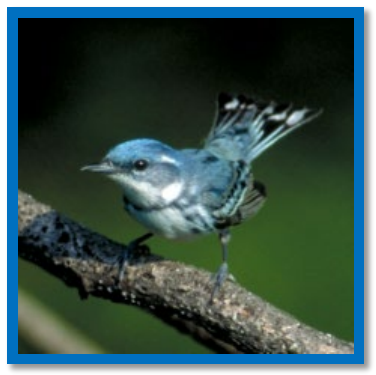}} &{\includegraphics[width=1.\linewidth]{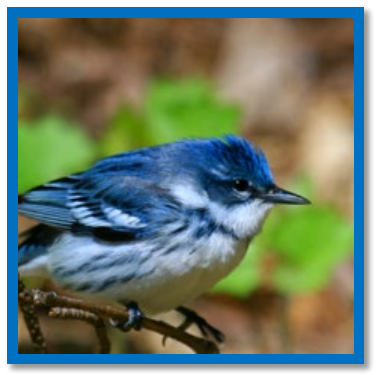}} &{\includegraphics[width=1.\linewidth]{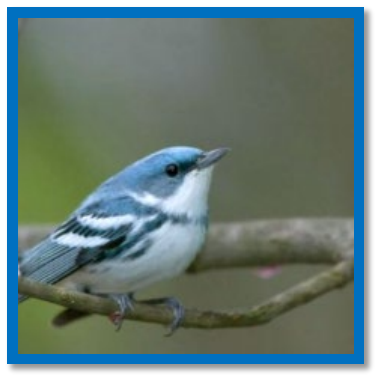}} &{\includegraphics[width=1.\linewidth]{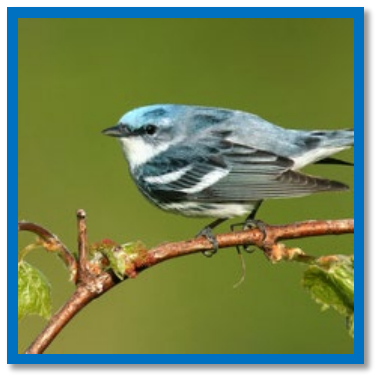}} &{\includegraphics[width=1.\linewidth]{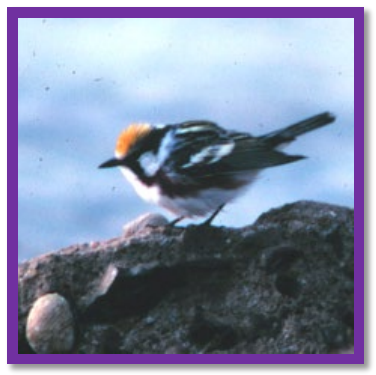}} &{\includegraphics[width=1.\linewidth]{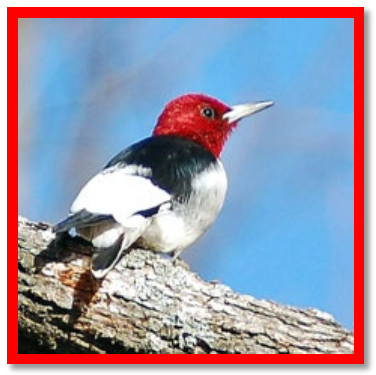}}\\
        \midrule[10pt]
        {\includegraphics[width=1.\linewidth]{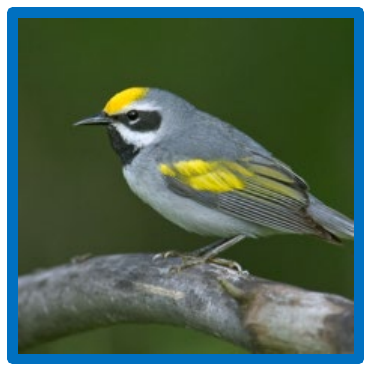}} &{\includegraphics[width=1.\linewidth]{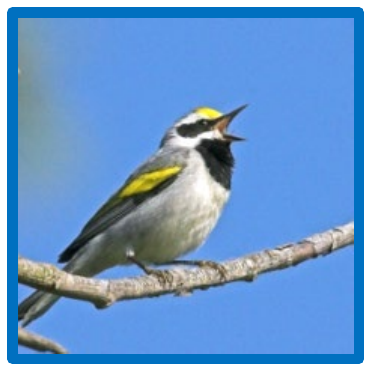}} &{\includegraphics[width=1.\linewidth]{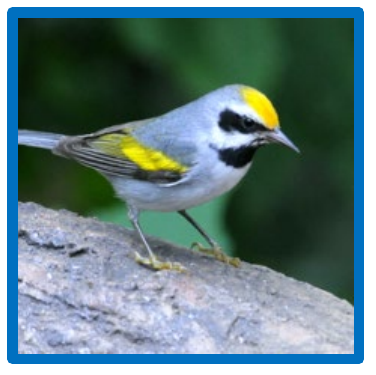}} &{\includegraphics[width=1.\linewidth]{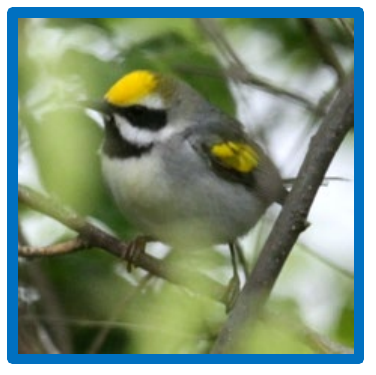}} &{\includegraphics[width=1.\linewidth]{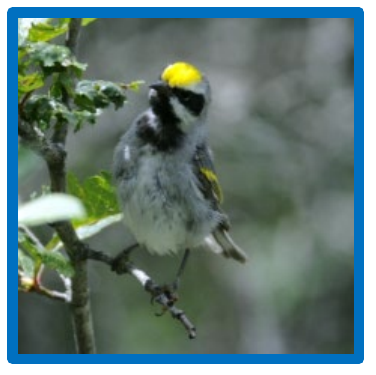}} &         {\includegraphics[width=1.\linewidth]{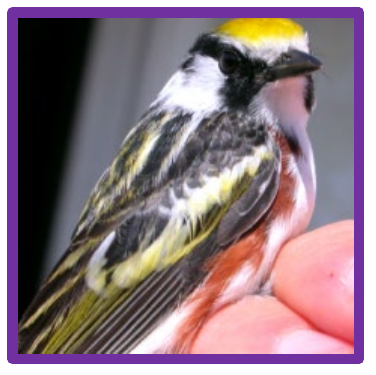}} & {\includegraphics[width=1.\linewidth]{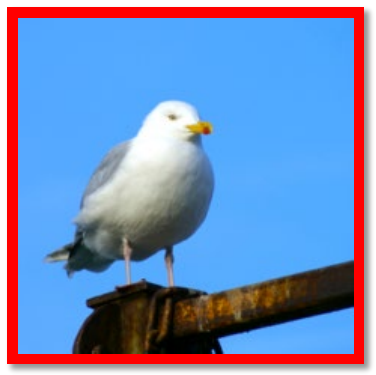}}\\
        \midrule[10pt]
        {\includegraphics[width=1.\linewidth]{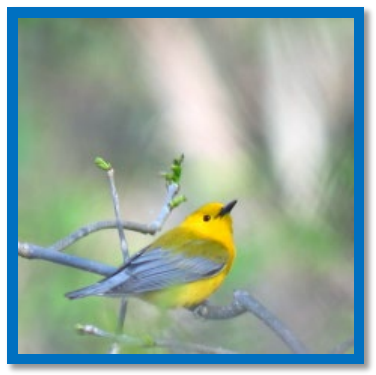}} &{\includegraphics[width=1.\linewidth]{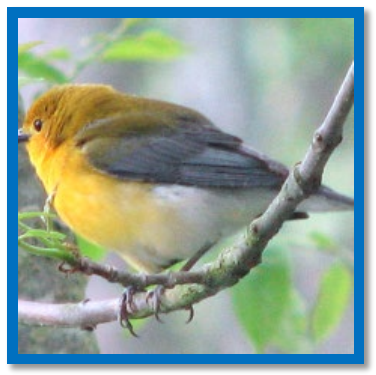}} &{\includegraphics[width=1.\linewidth]{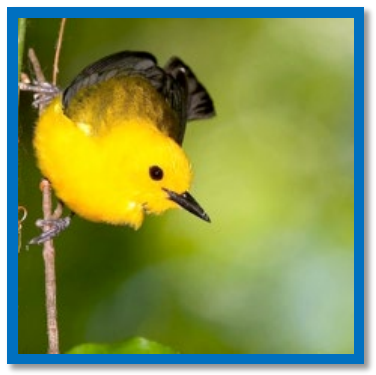}} &{\includegraphics[width=1.\linewidth]{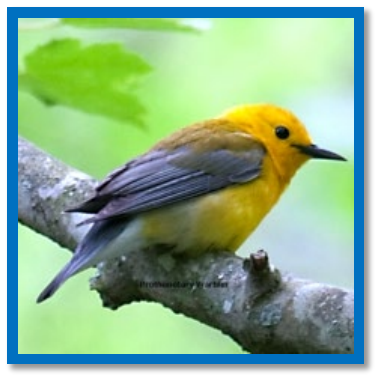}} &{\includegraphics[width=1.\linewidth]{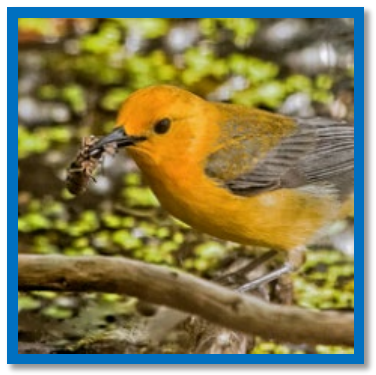}} &         {\includegraphics[width=1.\linewidth]{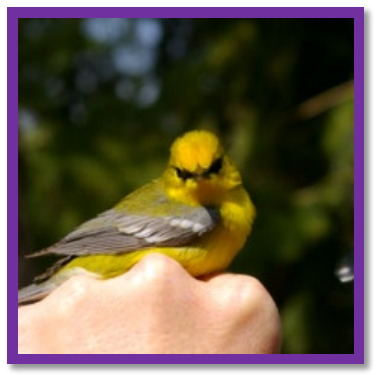}} & {\includegraphics[width=1.\linewidth]{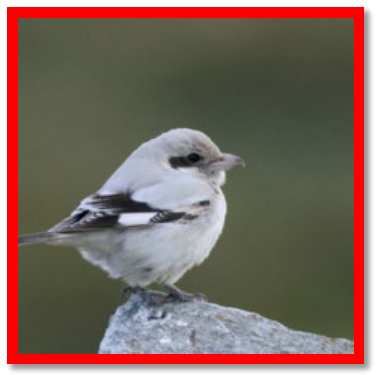}}\\
        \midrule[10pt]
        {\includegraphics[width=1.\linewidth]{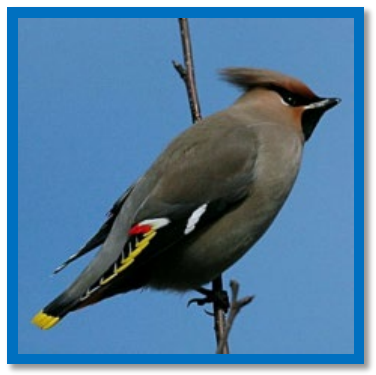}} &{\includegraphics[width=1.\linewidth]{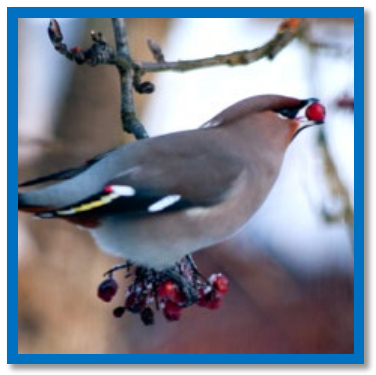}} &{\includegraphics[width=1.\linewidth]{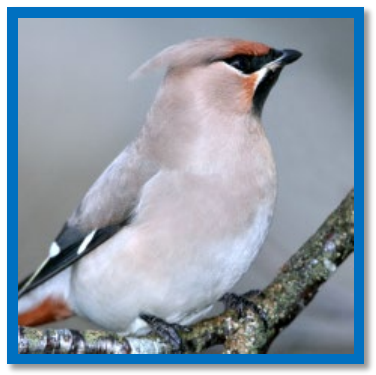}} &{\includegraphics[width=1.\linewidth]{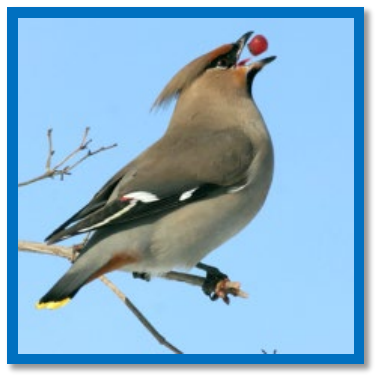}} &{\includegraphics[width=1.\linewidth]{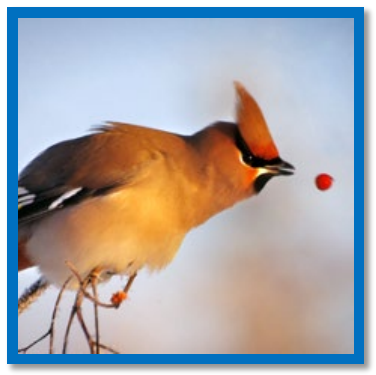}} &         {\includegraphics[width=1.\linewidth]{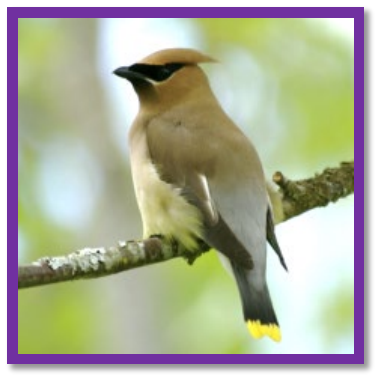}} & {\includegraphics[width=1.\linewidth]{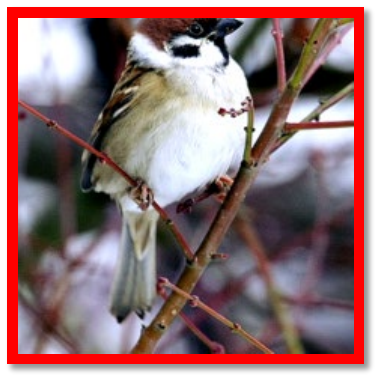}}\\
        \midrule[10pt]
        {\includegraphics[width=1.\linewidth]{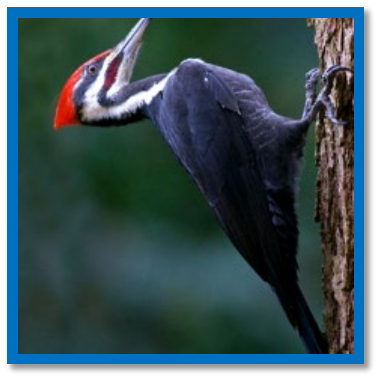}} &{\includegraphics[width=1.\linewidth]{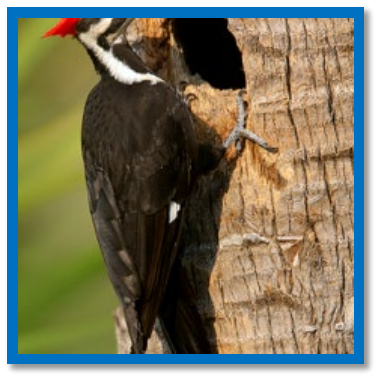}} &{\includegraphics[width=1.\linewidth]{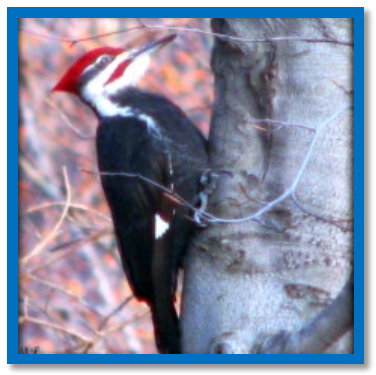}} &{\includegraphics[width=1.\linewidth]{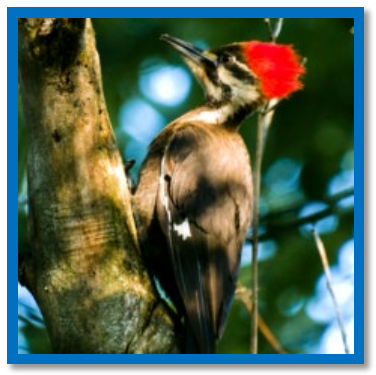}} &{\includegraphics[width=1.\linewidth]{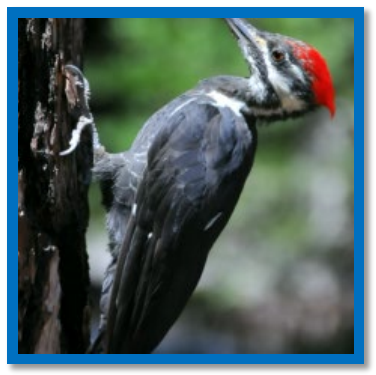}} &         {\includegraphics[width=1.\linewidth]{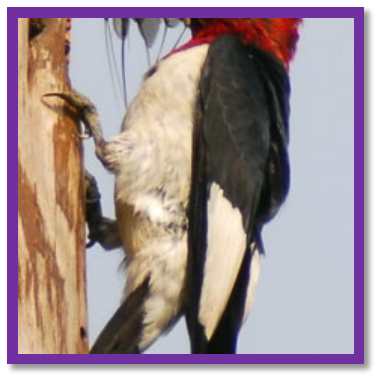}} & {\includegraphics[width=1.\linewidth]{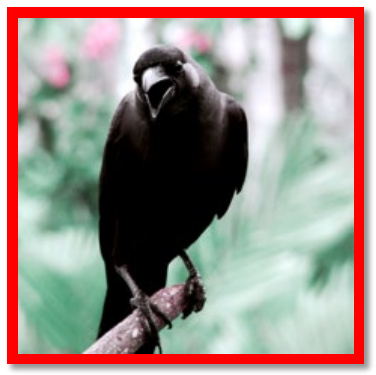}}\\
        \midrule[10pt]
        {\includegraphics[width=1.\linewidth]{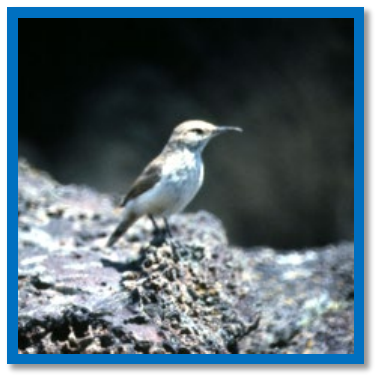}} &{\includegraphics[width=1.\linewidth]{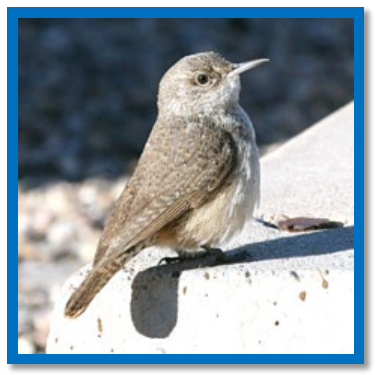}} &{\includegraphics[width=1.\linewidth]{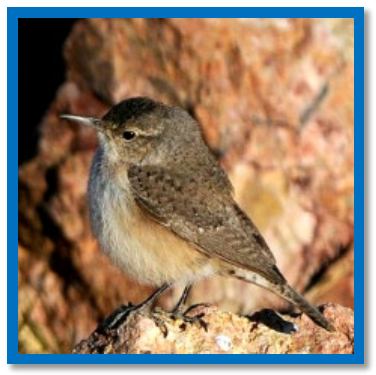}} &{\includegraphics[width=1.\linewidth]{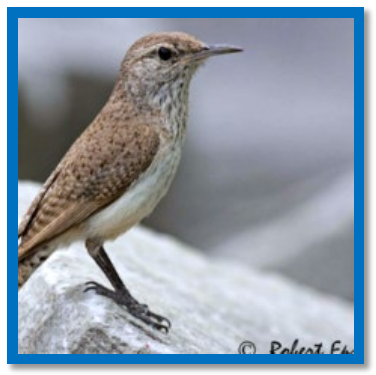}} &{\includegraphics[width=1.\linewidth]{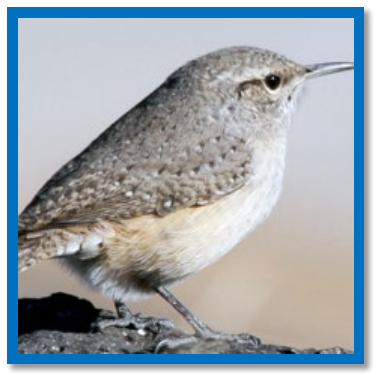}} &         {\includegraphics[width=1.\linewidth]{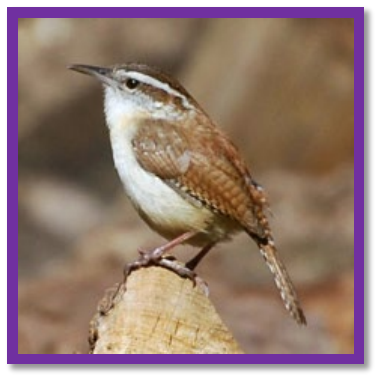}} & {\includegraphics[width=1.\linewidth]{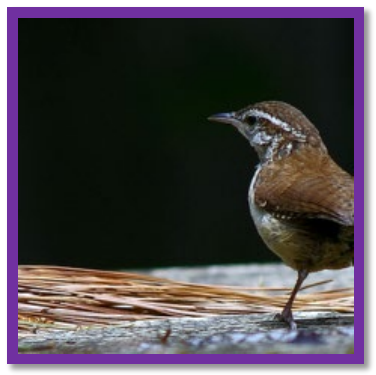}}\\
        \midrule[10pt]
        {\includegraphics[width=1.\linewidth]{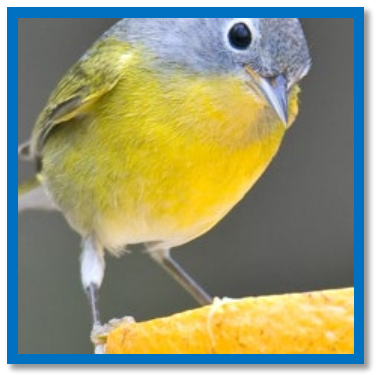}} &{\includegraphics[width=1.\linewidth]{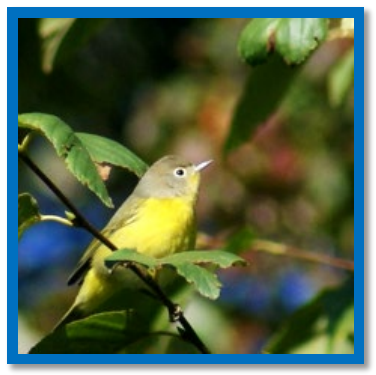}} &{\includegraphics[width=1.\linewidth]{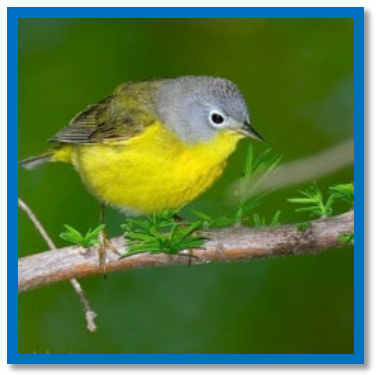}} &{\includegraphics[width=1.\linewidth]{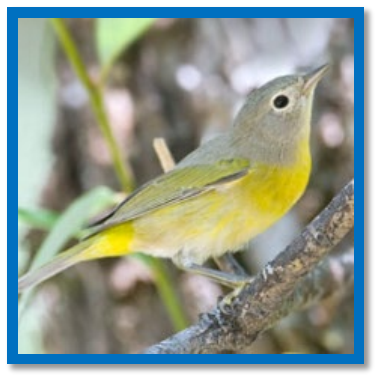}} &{\includegraphics[width=1.\linewidth]{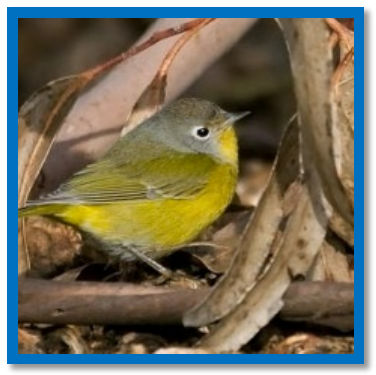}} &         {\includegraphics[width=1.\linewidth]{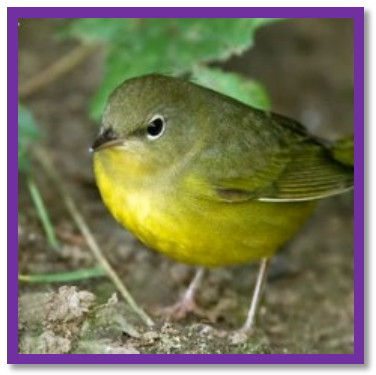}} & {\includegraphics[width=1.\linewidth]{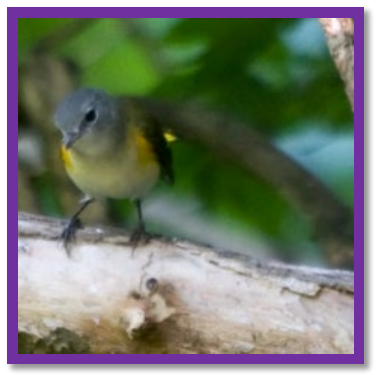}}\\        
    \end{tabular}}
    \caption{Qualitative experiment results of the proposed method. The evaluation is conducted using the CUB-200 dataset on ResNet-18, and the images belong to the novel categories and are also clustered in the novel categories. The first five columns with blue boxes denote well-clustered examples. The last two columns represent failed prediction results, including example images with purple boxes denoting hard negatives and those with red boxes indicating incorrect categorization.}
    \label{fig:experiment_quality_novel}
\end{figure*}

\begin{figure*}[t]
    \centering
    \small
    \resizebox{1\linewidth}{!}{
    \setlength{\tabcolsep}{1.pt}
    \begin{tabular}{ccccccc}
        {\includegraphics[width=1.\linewidth]{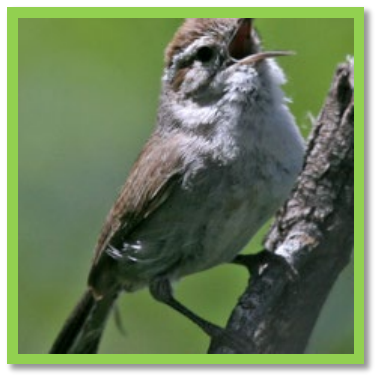}} &{\includegraphics[width=1.\linewidth]{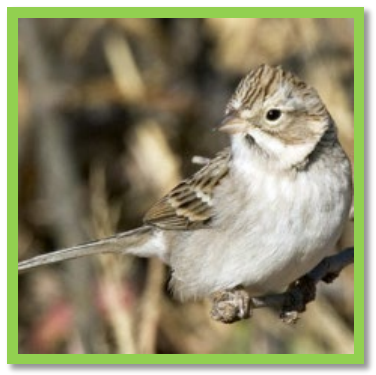}} &{\includegraphics[width=1.\linewidth]{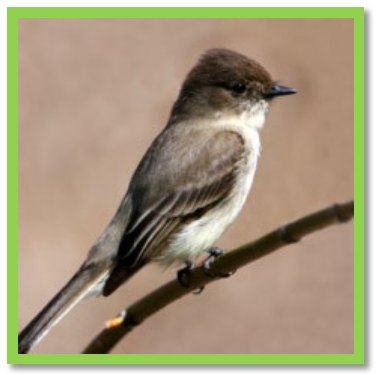}} &{\includegraphics[width=1.\linewidth]{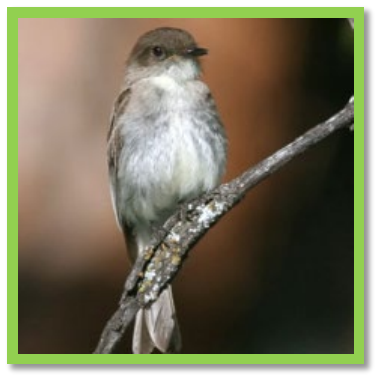}} &{\includegraphics[width=1.\linewidth]{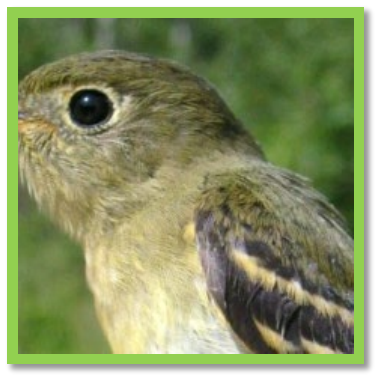}} &{\includegraphics[width=1.\linewidth]{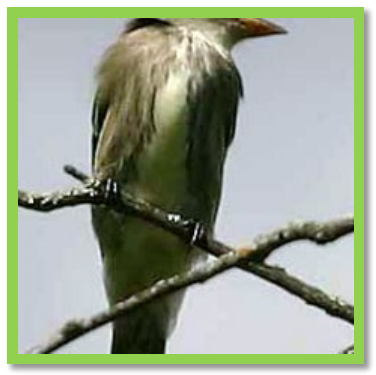}} &{\includegraphics[width=1.\linewidth]{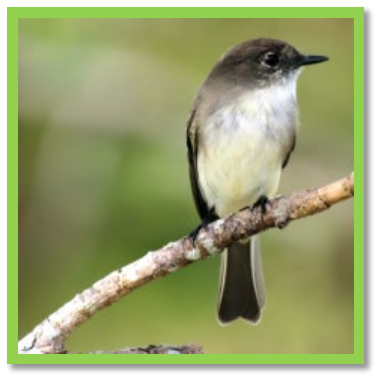}}\\
        \midrule[10pt]
        {\includegraphics[width=1.\linewidth]{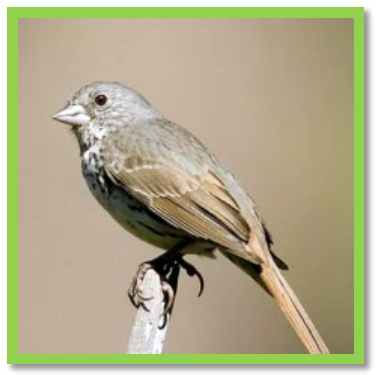}} &{\includegraphics[width=1.\linewidth]{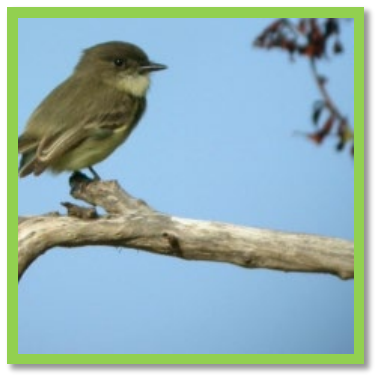}} &{\includegraphics[width=1.\linewidth]{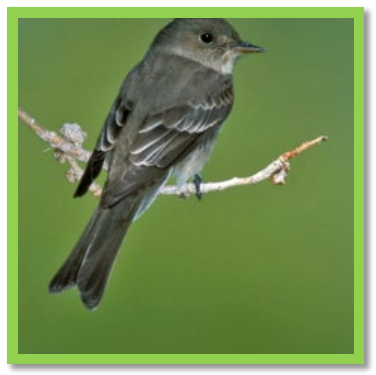}} &{\includegraphics[width=1.\linewidth]{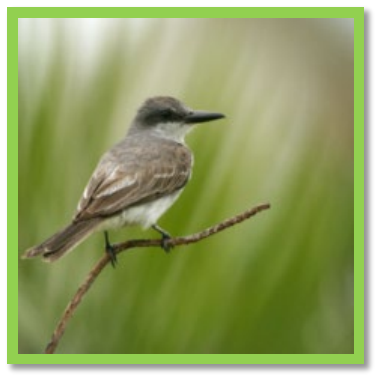}} &{\includegraphics[width=1.\linewidth]{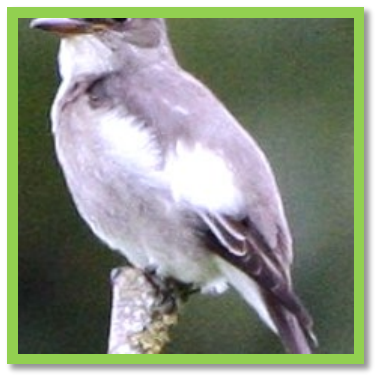}} &         {\includegraphics[width=1.\linewidth]{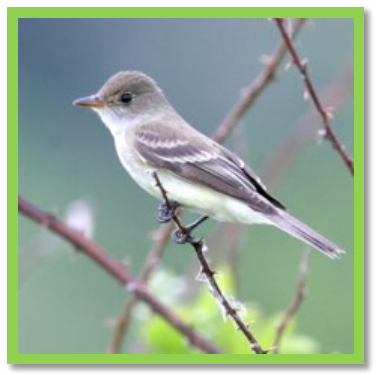}} & {\includegraphics[width=1.\linewidth]{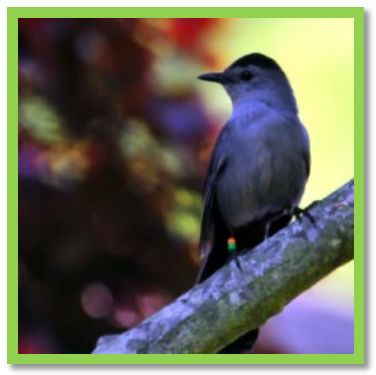}}\\
        \midrule[10pt]
        {\includegraphics[width=1.\linewidth]{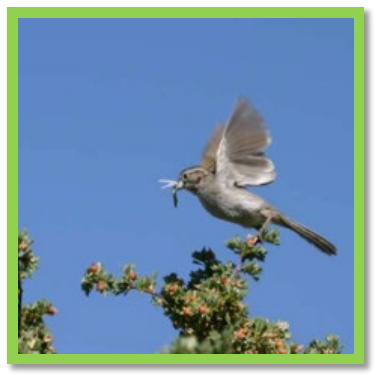}} &{\includegraphics[width=1.\linewidth]{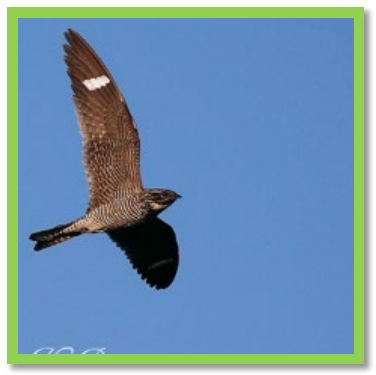}} &{\includegraphics[width=1.\linewidth]{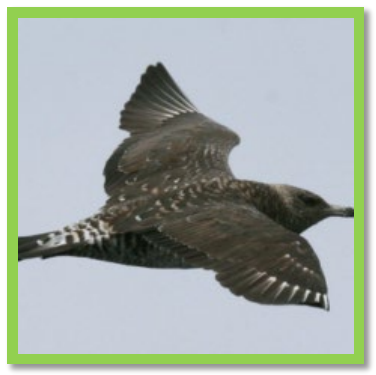}} &{\includegraphics[width=1.\linewidth]{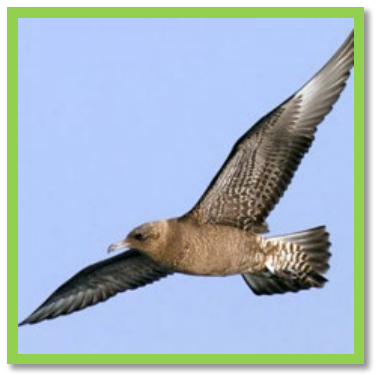}} &{\includegraphics[width=1.\linewidth]{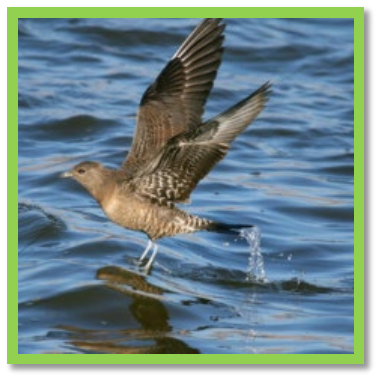}} &         {\includegraphics[width=1.\linewidth]{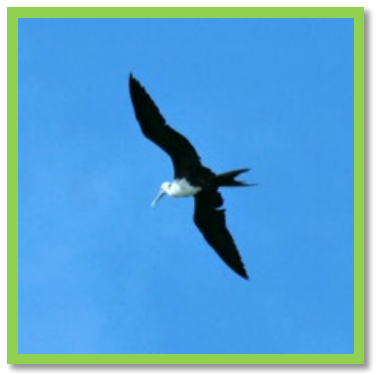}} & {\includegraphics[width=1.\linewidth]{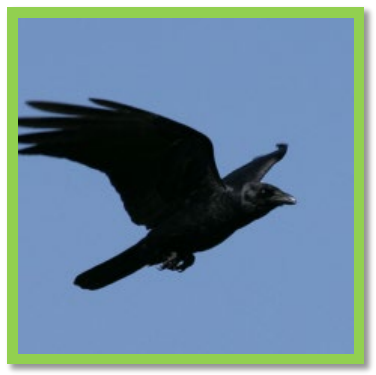}}\\
        \midrule[10pt]
        {\includegraphics[width=1.\linewidth]{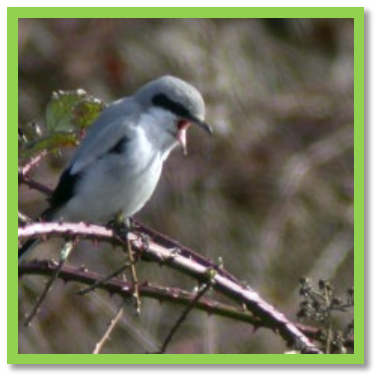}} &{\includegraphics[width=1.\linewidth]{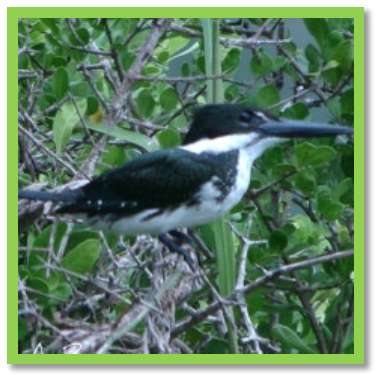}} &{\includegraphics[width=1.\linewidth]{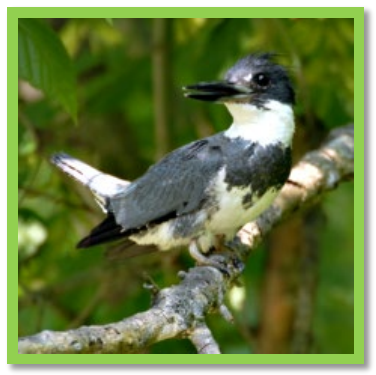}} &{\includegraphics[width=1.\linewidth]{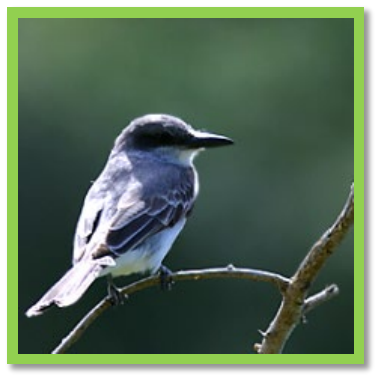}} &{\includegraphics[width=1.\linewidth]{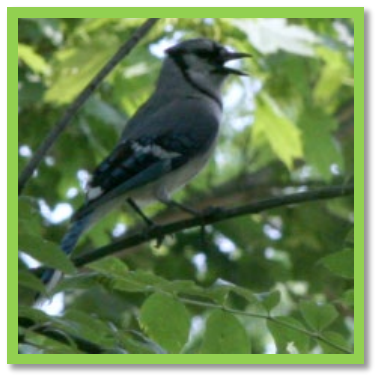}} &         {\includegraphics[width=1.\linewidth]{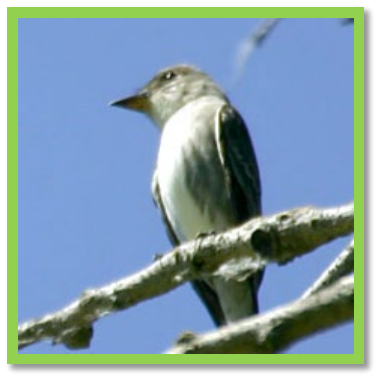}} & {\includegraphics[width=1.\linewidth]{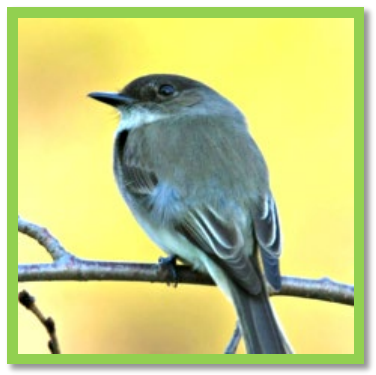}}\\
        \midrule[10pt]
        {\includegraphics[width=1.\linewidth]{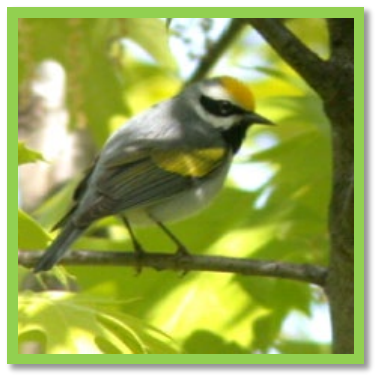}} &{\includegraphics[width=1.\linewidth]{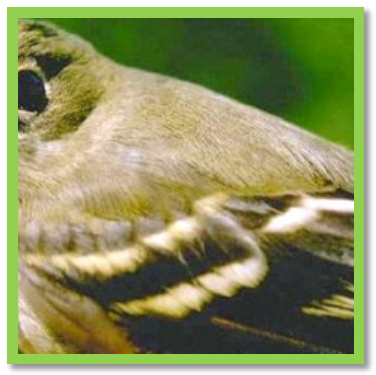}} &{\includegraphics[width=1.\linewidth]{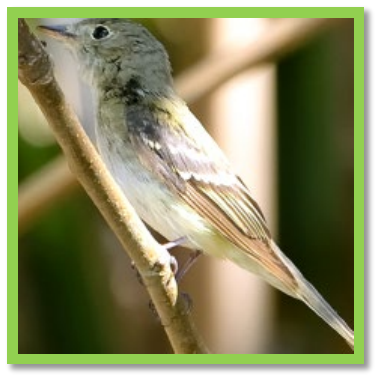}} &{\includegraphics[width=1.\linewidth]{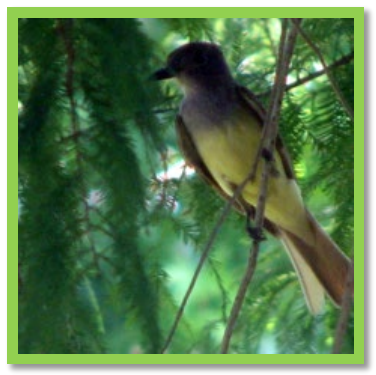}} &{\includegraphics[width=1.\linewidth]{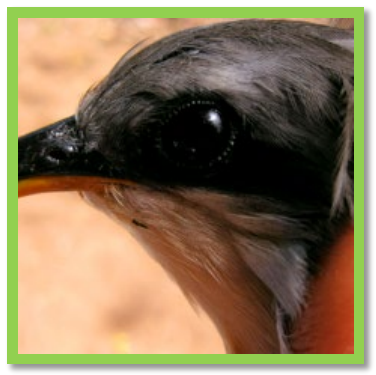}} &         {\includegraphics[width=1.\linewidth]{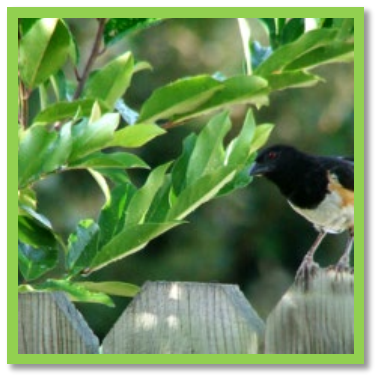}} & {\includegraphics[width=1.\linewidth]{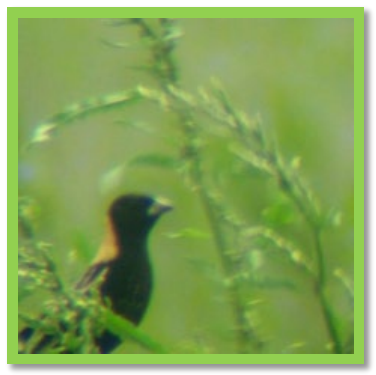}}\\
        \midrule[10pt]
        {\includegraphics[width=1.\linewidth]{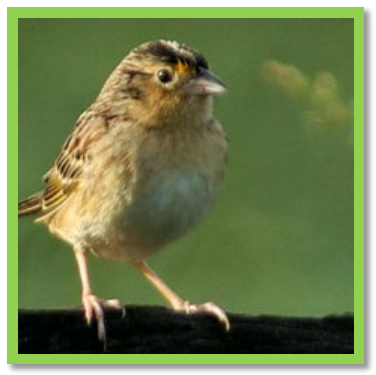}} &{\includegraphics[width=1.\linewidth]{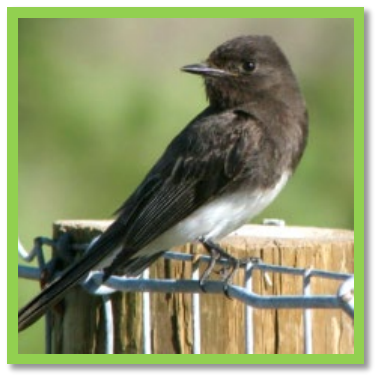}} &{\includegraphics[width=1.\linewidth]{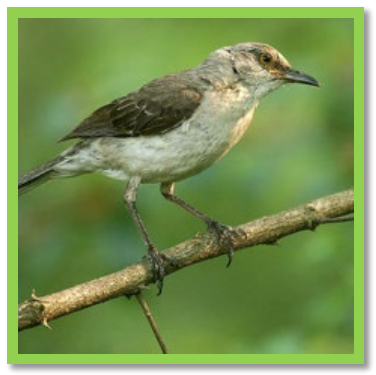}} &{\includegraphics[width=1.\linewidth]{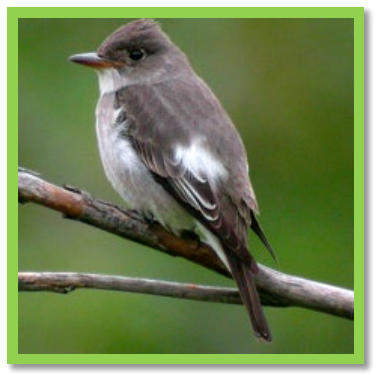}} &{\includegraphics[width=1.\linewidth]{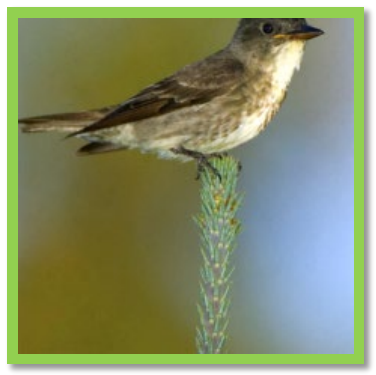}} &         {\includegraphics[width=1.\linewidth]{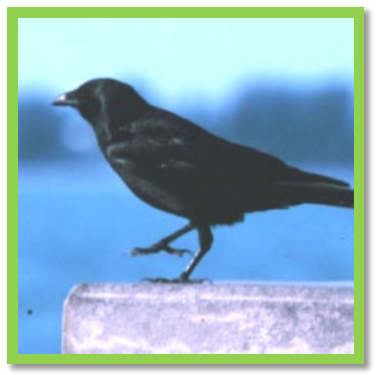}} & {\includegraphics[width=1.\linewidth]{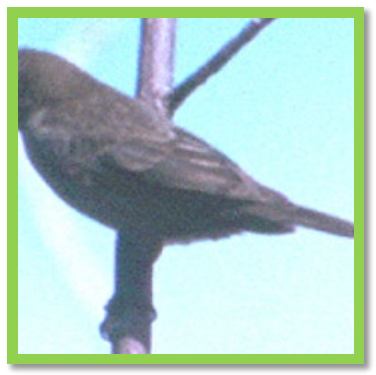}}\\
        \midrule[10pt]
        {\includegraphics[width=1.\linewidth]{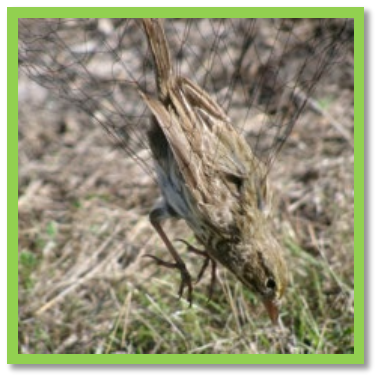}} &{\includegraphics[width=1.\linewidth]{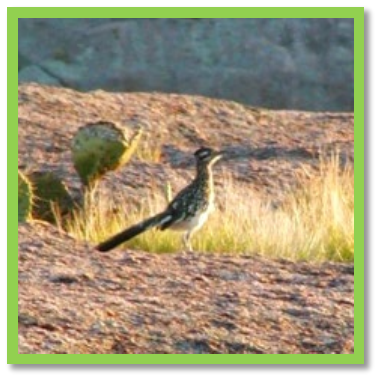}} &{\includegraphics[width=1.\linewidth]{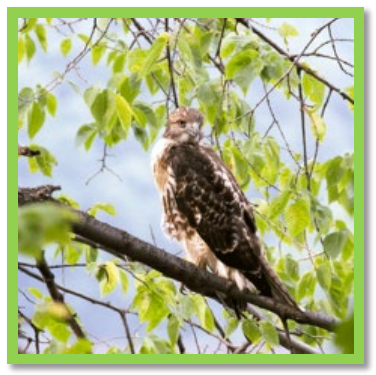}} &{\includegraphics[width=1.\linewidth]{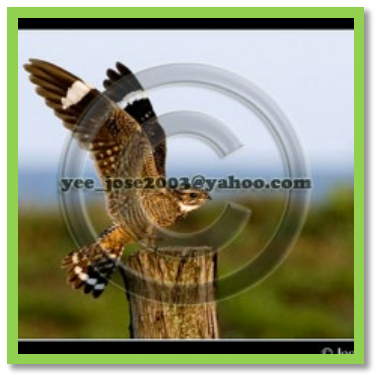}} &{\includegraphics[width=1.\linewidth]{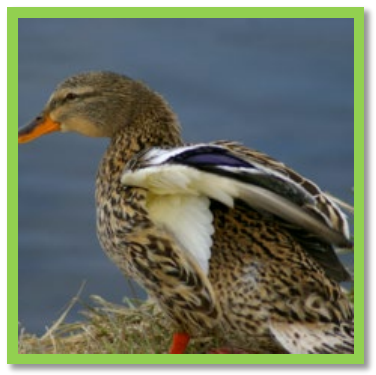}} &         {\includegraphics[width=1.\linewidth]{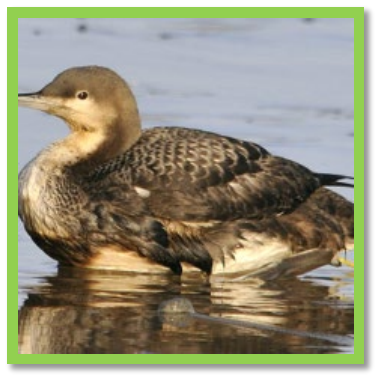}} & {\includegraphics[width=1.\linewidth]{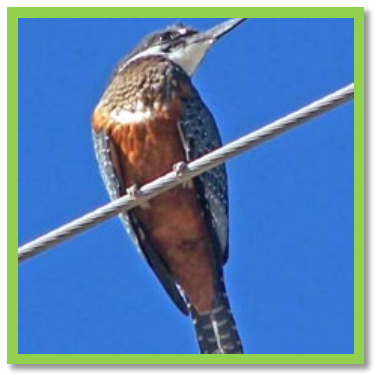}}\\          
    \end{tabular}}
    \caption{Qualitative experiment results of the proposed method. The evaluation is conducted using the CUB-200 dataset on ResNet-18. The images on each row have been categorized into the same classes, but it is not true. Nevertheless, the images on each row are hard to recognize the specific spices without the knowledge of experts. For that reason, we treat the images are hard negative samples.}
    \label{fig:experiment_quality_mingled}
\end{figure*}

To evaluate the proposed method qualitatively, we clustered the evaluation dataset using the CUB-200 dataset. 
As shown in~\cref{fig:experiment_quality_old}, our method well-discovered old categories and clustered them correctly. Each row is clustered into the same category, and the classes are old categories on the evaluation dataset. The left five columns are well-clustered, while the last two are not. The sixth-column images are still reasonable, but the last-column images are the worst cases. \cref{fig:experiment_quality_novel} depicted the clustered results of the novel categories discovery. Like the former result, each row image belongs to the same category. In contrast, the \cref{fig:experiment_quality_mingled} presents failure cases. The images on each row indicated that they were clustered into the same class. However, there are no images with the same label. Nevertheless, the images on each row have similar features, such as the colors of wings and feathers, and the behaviors. In this sense, we are hard to recognize that the categorized results failed without the specialized knowledge of the birds' species.


\end{document}